\documentclass[10pt,twocolumn,letterpaper]{article}

\usepackage{cvpr}
\usepackage{algorithm}
\usepackage{algorithmic}

\definecolor{cvprblue}{rgb}{0.21,0.49,0.74}
\usepackage[pagebackref,breaklinks,colorlinks,allcolors=cvprblue]{hyperref}

\begin{document}

\title{Generating on Generated: An Approach Towards Self-Evolving Diffusion Models}


\author{
Xulu Zhang\textsuperscript{1,2},
Xiaoyong Wei\textsuperscript{1},
Jinlin Wu\textsuperscript{2,3},
Jiaxin Wu\textsuperscript{1},
Zhaoxiang Zhang\textsuperscript{2,3,4},
Zhen Lei\textsuperscript{2,3,4},
Qing Li\textsuperscript{1} \\
}
\twocolumn[{
\maketitle

\begin{center}
    \textsuperscript{1}The Hong Kong Polytechnic University, Hong Kong \\
    \textsuperscript{2}CAIR, HKISI, CAS, Beijing \\
    \textsuperscript{3}UCAS, Beijing \\
    \textsuperscript{4}CASIA, Beijing
    \captionsetup{type=figure}
    \vspace{2pt}
    \includegraphics[width=.95\textwidth]{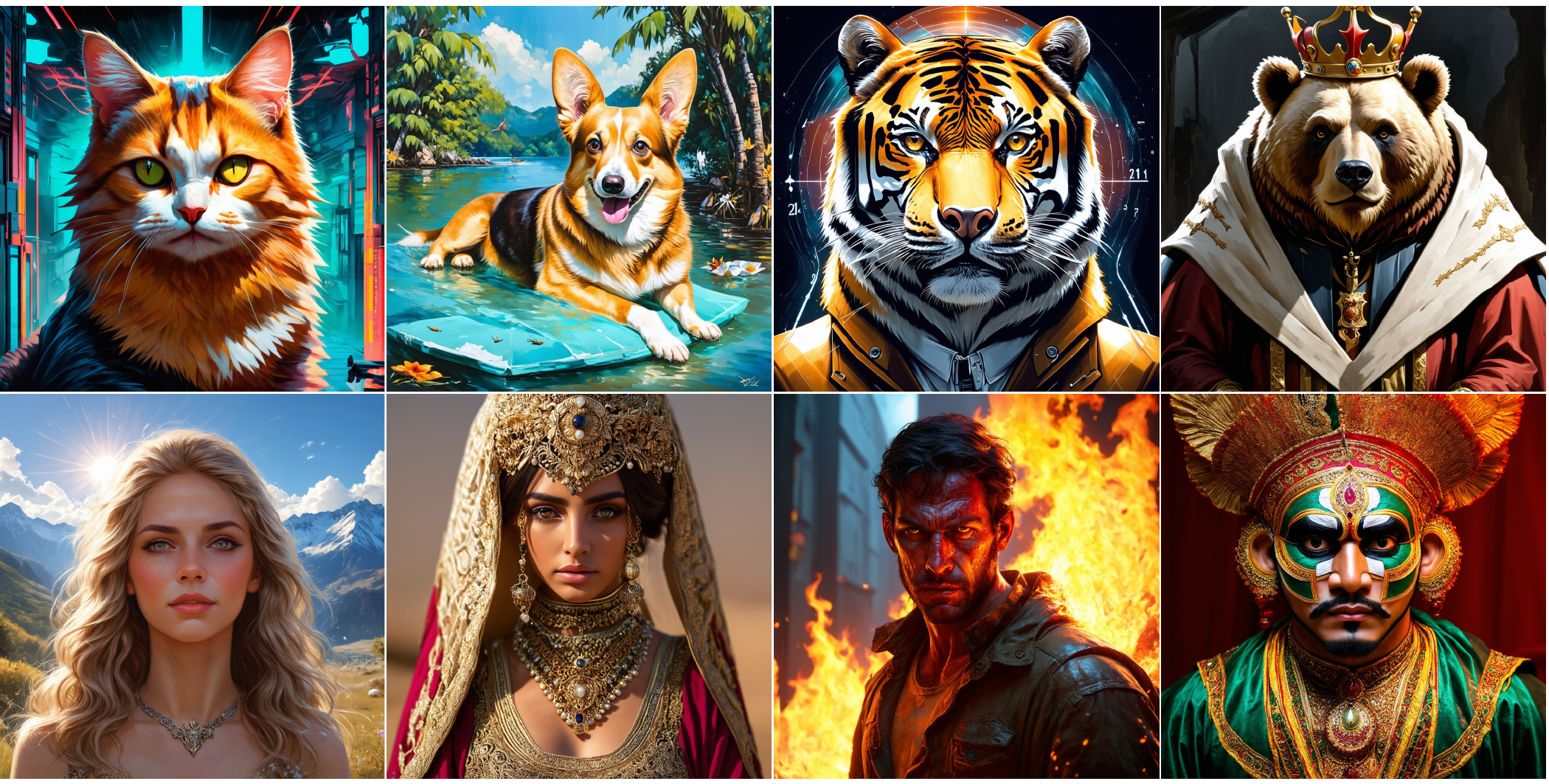}
    \captionof{figure}{We introduce RSIDiff, a novel approach that enhances the performance of diffusion models through recursive self-training. By iteratively refining the model with its own generated data, RSIDiff produces images of noteworthy aesthetic quality.}
\end{center}
}]

\begin{abstract}
\vskip -0.1in
    Recursive Self-Improvement (RSI) enables intelligence systems to autonomously refine their capabilities. This paper explores the application of RSI in text-to-image diffusion models, addressing the challenge of training collapse caused by synthetic data. We identify two key factors contributing to this collapse: the lack of perceptual alignment and the accumulation of generative hallucinations.
    To mitigate these issues, we propose three strategies: (1) a prompt construction and filtering pipeline designed to facilitate the generation of perceptual aligned data, (2) a preference sampling method to identify human-preferred samples and filter out generative hallucinations, and (3) a distribution-based weighting scheme to penalize selected samples with hallucinatory errors. Our extensive experiments validate the effectiveness of these approaches.
    %
\end{abstract}

\section{Introduction}

The concept of self-evolving generative models \cite{tao2024survey} holds increasing significance within the realm of machine learning. 
Recently, Recursive Self-Improvement methods \cite{schmidhuber2003godel,yampolskiy2015seed,nivel2013bounded,steunebrink2016growing} have garnered substantial attention. 
RSI strives for superintelligence by empowering artificial general intelligence systems to autonomously enhance their own capabilities.
This approach has been exemplified by systems like AlphaGo \cite{silver2016mastering}, which utilize self-play to significantly improve decision-making capabilities, ultimately defeating human champions. 
Moreover, there has been growing interest in applying RSI to language models (LMs) \cite{zelikman2022star,gou2023critic,yuan2024self,chen2024self}. 
By incorporating RSI, LMs can iteratively improve their linguistic understanding, leading to more nuanced and contextually aware interactions. 

Recently, diffusion models have emerged as a prominent area of research. 
By employing denoising strategies and pre-training on large-scale datasets, advanced methods, such as DALL-E 2 \cite{ramesh2022hierarchical} and Stable Diffusion \cite{rombach2022high}, have shown a remarkable capacity for generating compelling and diverse samples. 
However, the scarcity of reliable data and growing privacy concerns present significant challenges to the ongoing advancement of these models. 
%
In this context, RSI becomes a promising solution, where models could leverage synthetic data for continuous self-evolving while addressing both data availability and privacy issues.

\begin{figure}[t]
\centering
\centerline{\includegraphics[width=0.93\linewidth]{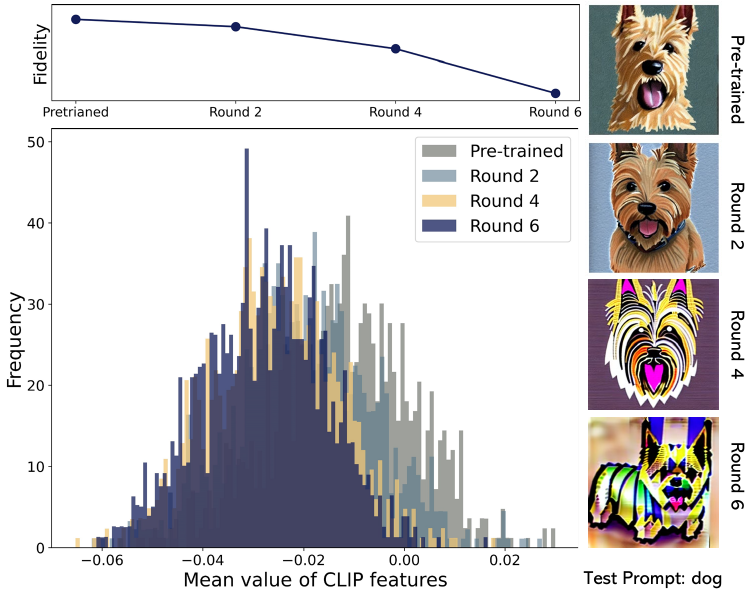}}
\caption{\textbf{Degeneration in RSI.} We observe a severe domain shift and decline in image fidelity when fine-tuning diffusion models with self-generated data. The diffusion model gradually loses the ability to generate fine-grained details.}
\label{fg:domain_shift}
\vskip -0.1in
\end{figure}

Despite its potential, implementing RSI in diffusion models through self-training presents technical challenges. 
%
The primary one is the training collapse \cite{shumailov2024ai,dohmatob2024strong,alemohammad2023self}, which frequently occurs during fine-tuning with randomly selected data.
This collapse results in the production of defective outputs, creating a detrimental feedback loop where poor results hinder further model refinement.
We also observe a similar outcome when training diffusion models with their own generated data. 
As illustrated in \cref{fg:domain_shift}, there is a noticeable increase in artifacts in the generated images.
%
The diffusion model gradually loses the ability to depict fine-grained details and can only generate the rough contour of the desired dog.
Additionally, we notice an evident distribution shift accompanying this image degradation.

In this paper, we attribute the model collapse to the lack of \textbf{perceptual alignment} and the compounding of \textbf{generative hallucinations}. 
Specifically, \textbf{perceptual alignment} refers to the ability of the model to generate outputs that closely match human preferences and expectations, such as accurately depicting the style and context of a given prompt.
%
However, due to the inherent randomness of diffusion models, the number of unsatisfactory samples generated often far exceeds that of human-preferred ones.
%
As a result, random sampling struggles to provide perceptual-aligned information for self-improvement.
Conversely, \textbf{generative hallucinations} frequently exist in out-of-distribution samples, such as ``dog with five legs'' or ``human with seven fingers''. These flawed images are included in the training data through random sampling and poison the model training. This sustained accumulation eventually results in the domain shift and degraded generation quality.
Therefore, it is essential to explore strategies that avoid generative hallucinations in generated data while enhancing perceptual alignment to promote continuous improvement.

\begin{figure*}[t]
\begin{center}
\centerline{\includegraphics[width=0.93\linewidth]{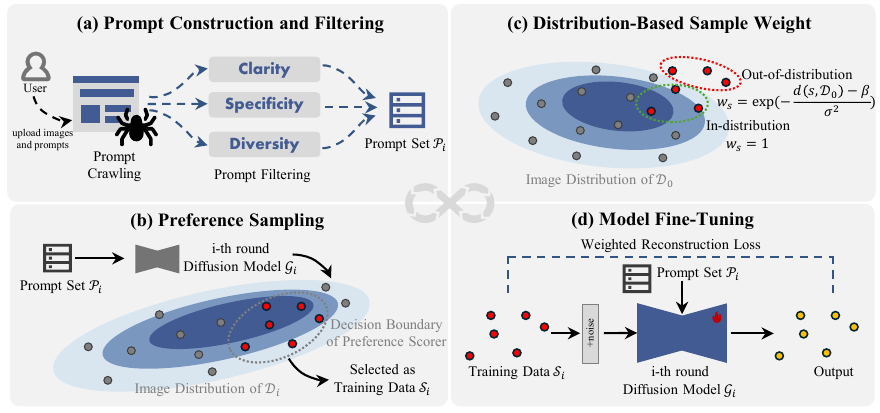}}
\caption{\textbf{Framework of RSIDiff.} (a) We crawl Prompts from user-active image synthesis website and filter them based on clarity, specificity, and diversity; (b) We employ preference sampling, which utilizes automatic metrics to identify human-preferred images; (c) We use the distribution-based weighting strategy to penalize out-of-distribution samples; and (d) We fine-tune the diffusion model with the selected samples and start a new training round.}
\label{fg:framework}
\end{center}
\vskip -0.2in
\end{figure*}

To address these challenges, we propose that the diffusion model needs a mechanism to control its perceptual alignment and reduce the generative hallucinations when performing RSI. 
To this end, we review the training process of text-to-image diffusion models and find that the quality of the text prompt and its corresponding generated image highly impact the model evolution.
Driven by this intuition, we introduce a mechanism focusing on both data quality and distribution to construct better prompt-(synthetic)image pairs to control the self-evolving process. 
For data quality, we design a pipeline for high-quality prompt construction. This process aims to provide clear, specific, and diverse prompts to improve the perceptual alignment of synthetic images. 
%
In addition, we introduce a preference sampling strategy aimed at selectively curating the synthetic data with fewer hallucinations and better human preference scores.
%
For the data distribution, we propose a distribution-based weighting scheme to penalize the out-of-distribution samples. This minimizes the impact of potential hallucinatory errors.
%
%
The overview of the three proposed strategies are visualized in \cref{fg:framework} (a), (b), and (c), respectively. 
%
%
Through these methods, we aim to build a robust, effective, and controllable RSI process for diffusion models.

\textbf{Contribution.} 
1) We conduct a pilot study exploring the application of RSI in text-to-image diffusion models;
2) We propose three approaches designed to promote perceptual alignment while minimizing the impact of generative hallucinations;
3) We perform extensive experiments to validate the effectiveness of the proposed approaches.

\section{Related Work}
\subsection{Recursive Self-Improvement}
RSI was first introduced to solve harder problems through iterative refinement \cite{schmidhuber2003godel}.
A significant sub-area is the use of synthetic data for self-retraining, which has gained much discussion about its associated challenges.
For instance, \cite{shumailov2024ai} demonstrates that iterative training of models like OPT-125M \cite{zhang2022opt} with synthetic text can lead to severe distribution shifts.
Similarly, \cite{dohmatob2024strong} has found that even a small number of synthetic samples can cause significant large model collapse during training.
Meanwhile, \cite{alemohammad2023self} reports substantial degradation in the quality of generated samples when using synthetic images to train StyleGAN2 \cite{karras2020analyzing} and DDPM models \cite{ho2020denoising}.

Despite these challenges, recent work by Meta \cite{yuan2024self} has proven the feasibility of RSI with synthetic data. They propose a self-rewarding model that utilizes large language models (LLMs) to generate responses and corresponding rewards based on a generated prompt set. These sample pairs are then used for Direct Preference Optimization (DPO) \cite{rafailov2024direct}.
While these studies have effectively highlighted the challenges in RSI using synthetic data, they have not specifically discussed state-of-the-art (SoTA) text-to-image models.
In contrast, this paper aims to address these gaps by focusing on the application of RSI to SoTA diffusion models and providing practical solutions.

In addition to using synthetic data, there has been other exploration of self-improvement strategies. 
These methods primarily utilize self-critique mechanisms \cite{madaan2024self,chen2023teaching,welleckgenerating,han2024small,miao2023selfcheck}, where the model refines its response based on its own feedback. 
However, these dialogue-based methods cannot be directly applied in the context of text-to-image models.

\subsection{Human Preference Alignment}
In recent advancements of text-to-image models, a primary focus has been on enhancing alignment with human preferences. 
This typically requires extensive pre-training on high-quality image-text pairs \cite{betker2023improving,esser2024scaling}.
Another research branch focuses on fine-tuning pre-trained models to yield better results.
For example, \cite{dai2023emu} fine-tunes SDXL \cite{podell2023sdxl} with 2,000 high-quality samples, which are filtered from a 1.1 billion dataset.
Additionally, some methods leverage reinforcement learning during the tuning phase. 
For example, DDPO \cite{black2023training} and DPOK \cite{fan2024reinforcement} have trained a reward model to guide the optimization, while Diffusion-DPO \cite{wallace2024diffusion} and D3PO \cite{yang2024using} propose the implicit reward functions to fine-tune diffusion models directly on the preference dataset.
However, there remains limited exploration into the potential of using self-generated data for further improvement.

Meanwhile, a variety of human preference datasets and evaluation metrics have been developed. Pic-a-pic dataset \cite{kirstain2023pick} features 583,747 image pairs for binary comparison, and the HPD v2 dataset \cite{wu2023human} includes 798,090 human preference choices across 433,760 image pairs. 
The evaluation metrics for assessing human preferences are also advanced significantly, including PickScore \cite{kirstain2023pick}, Human Preference Score (HPS) \cite{wu2023humanhps}, HPS v2 \cite{wu2023human}, and image rewards \cite{xu2024imagereward}. 
These together provide a robust framework for evaluating how well generated content aligns with human expectations.

\begin{algorithm}[tb]
   \caption{RSIDiff Procedure}
   \label{alg:rsidiff_procedure}
\begin{algorithmic}
   \STATE {\bfseries Input:} Base model $\mathcal{G}_0$, prompt set $\mathcal{P}=\left \{ \mathcal{P}_0, \mathcal{P}_1, ...,\mathcal{P}_r \right \}$, total training round $r$.
   \STATE Initialize current training round $i=0$.
   \REPEAT
   \STATE Generate synthetic dataset $\mathcal{D}_i$ using prompt set $\mathcal{P}_i$ with the diffusion model $\mathcal{G}_i$.
   \STATE Apply preference sampling to obtain the synthetic training set $\mathcal{S}_i$ from dataset $\mathcal{D}_i$.
   \STATE Calculate the sample weights $\mathcal{W}_i$ of training set $\mathcal{S}_i$.
   \STATE Fine-tune the model $\mathcal{G}_i$ with weighted training set $(\mathcal{W}_i\circ\mathcal{S}_i,\mathcal{P}_i)$ and get the updated model $\mathcal{G}_{i+1}$.
   \STATE Update current training round $i=i+1$.
   \UNTIL{Round $i$ reaches the total training round $r$.}
\end{algorithmic}
\end{algorithm}

\section{Method}

\subsection{Problem Definition}
Given a base model $\mathcal{G}_0$, RSI aims to enhance the performance of target models through iterative refinement.
At each round $i$, the model generates a set of synthetic data $\mathcal{D}_i$ based on the current prompt set $\mathcal{P}_i$. This generated data serves as the primary input for the subsequent training phase within the RSI loop.
The fine-tuning process can be formulated as follows:
\begin{equation}
    \theta_{i+1}=f(\theta_i, \mathcal{D}_i, \mathcal{P}_i),
\end{equation}
where $\theta_i$ denotes the target model parameters at round $i$, and $f$ is a function that updates the model based on the received data and associated loss feedback.

In this paper, we focus on the RSI of diffusion models. The diffusion process consists of two main steps: a noise-adding step in the forward process and a denoising step in the reverse process. During the forward process, Gaussian noise is incrementally added to the input data $x_0$, resulting in a noisy latent representation given by

\begin{equation}
    x_t=\sqrt{\alpha_t} x_{t-1}+\sqrt{1-\alpha_t}\epsilon_t, 
\end{equation}
where $\alpha_t$ controls the variance of the added noise $\epsilon_t$. The reverse process, denoted as $\hat{x_{\theta}}(x_{t}, t, c)$, aims to denoise the noisy latent $x_t$ based on the current timestep $t$ and the prompt condition $c$.

Consequently, the update function $f$ can be written as:

\begin{equation}
\label{eq:update}
    f(\theta, \mathcal{D}, \mathcal{P}) = \min_\theta\frac{1}{|\mathcal{D}|}\sum_{x_0 \in \mathcal{D}, c \in \mathcal{P}} \mathbb{E}\left [ \lambda_{t}||\hat{x_{\theta}}(x_{t}, t, c)- x_0) ||^ {2} \right ]
\end{equation}
where $\lambda_{t}$ is a time-dependent weighting factor. The goal is to minimize the difference between the denoised output and the original data, thereby refining the model's capability with synthetic data.

\begin{figure}[t]
\begin{center}
\centerline{\includegraphics[width=0.98\linewidth]{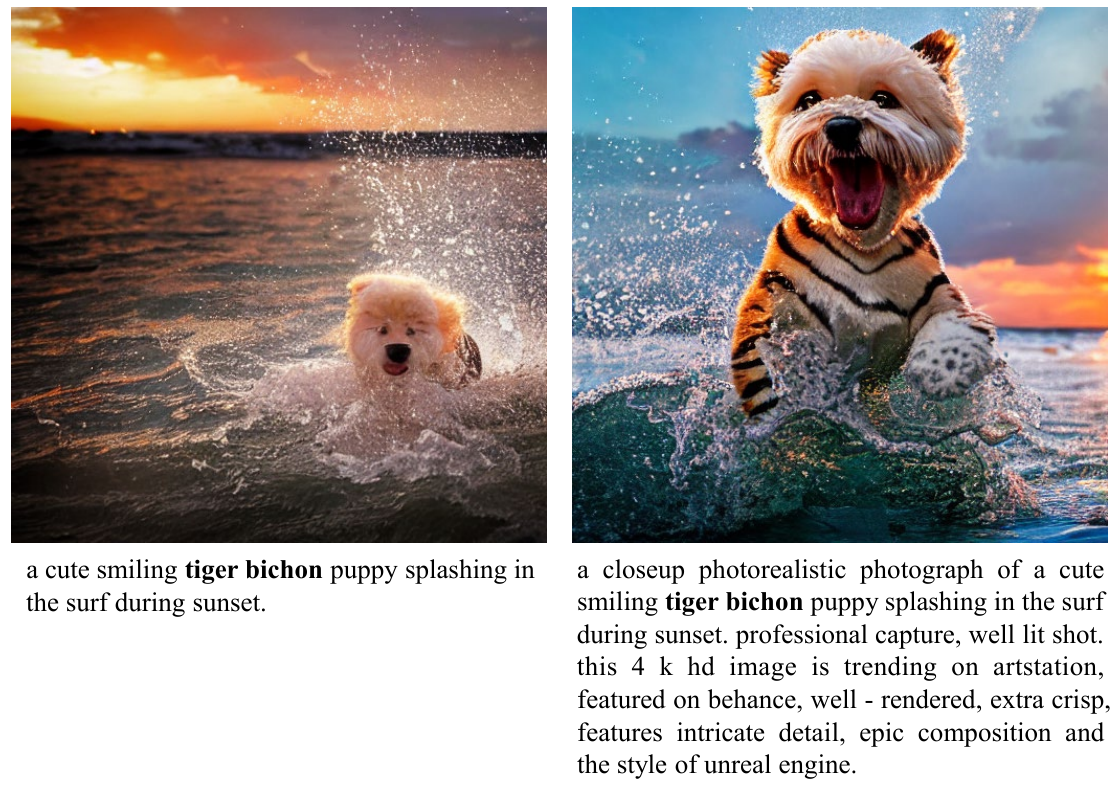}}
\caption{Examples generated by SD v1.4 with a simple prompt (left) and our filtered prompt (right). The latter produces a more visually appealing image.}
\label{fg:prompt_samples}
\end{center}
\vskip -0.2in
\end{figure}

\subsection{Better Prompts}
Recently, text-to-image models have shown significant improvements when trained with highly descriptive image captions \cite{betker2023improving,esser2024scaling}. 
In the context of RSI, the prompts used for synthetic image generation not only guide the model in producing images but also serve as captions in subsequent training iterations.
Additionally, due to the random nature of diffusion models, well-designed prompts can significantly increase the likelihood of generating outputs that align with human preferences. 
As illustrated in \cref{fg:prompt_samples}, we compare generations from two types of prompts and carefully crafted prompts yield visually appealing results.

To acquire such high-quality prompts, we utilize a dataset of 172k prompts collected from the Lexica website, where users share preferred synthetic images and corresponding prompts. 
%
%
To further optimize the efficacy of these prompts, we propose a structured filtering approach that focuses on \textbf{clarity}, \textbf{specificity}, and \textbf{diversity}. 

\textbf{Clarity} reduces ambiguity in the prompts to ensure better embedding representations in the model. We achieve this by designing instructions to ask Llama 3 \cite{dubey2024llama} for filtering.

\textbf{Specificity} includes detailed attributes and contexts that accurately represent the desired image, which is also implemented through Llama 3 filtering.

\textbf{Diversity} guarantees a wide range of scenarios in the prompts to promote a comprehensive learning procedure. We implement this by using K-means clustering \cite{ahmed2020k} to categorize prompt embeddings into distinct groups. Prompts that are closest to the cluster centers are selected.

Through this human-aligned prompt construction process and quality-oriented filtering strategy, we extract 40k prompts that significantly improve the perceptual alignment of synthetic images and benefit the following methods.

\subsection{Preference Sampling}
Recent research has shown that using random sampling methods can lead to model collapse.
In this paper, we explain this phenomenon by analyzing the two impacts from generated samples: \textbf{perceptual alignment} and \textbf{generative hallucinations}. 
\textbf{Perceptual alignment} provides the human-preferred information exhibited in the generated samples that enhances the model's abilities.
In contrast, \textbf{generative hallucinations} are negative information arising from defective data that hinder model training.
As we have discussed, random sampling fails to find valuable samples for model optimization while simultaneously introducing generative hallucinations. 
This leads to the accumulation of errors through iterative training processes and causes severe distribution shifts and overall performance degradation.

To address these challenges, this paper proposes a preference sampling strategy to identify preferred data and filter out flawed samples.
%
Specifically, we employ automated metrics to evaluate the preference of the synthetic data. These metrics encompass various aspects, including the alignment between text prompts and corresponding generated images \cite{radford2021learning}, aesthetic quality \cite{schuhmann2022laion}, and overall human preference scores \cite{wu2023human}. 
This samples a training subset $\mathcal{S}_i$ from the generated data $\mathcal{D}_i$ at round $i$. The \cref{eq:update} will be updated as follows:

\begin{equation}
    f(\theta, \mathcal{S}, \mathcal{P}) = \min_\theta\frac{1}{|\mathcal{S}|}\sum_{x_0 \in \mathcal{S}, c \in \mathcal{P}} \mathbb{E}\left [ \lambda_{t}||\hat{x_{\theta}}(x_{t}, t, c)- x_0) ||^ {2} \right ]
\end{equation}

This ensures that the filtered dataset exhibits a strong alignment with human preferences and minimizes undesired generative hallucinations.

\subsection{Distribution-based Sample Weight}
While preference sampling selects the generated data that aligns well with human preferences, hallucinations may still exist. 
To maintain continuous improvement, it is important to evaluate the contribution of each sample to model optimization. 
As mentioned in \cref{fg:domain_shift}, we observe that the accumulation of out-of-distribution hallucinations can lead to significant domain shifts.
To address this issue, we propose a distribution-based weighting scheme to assess the potential errors in the selected samples.
This scheme uses the samples $\mathcal{D}_0$ generated by the base model $\mathcal{G}_0$ as references to evaluate the distribution shift of selected samples $\mathcal{S}_i$.
In this scheme, samples that are located within the main distribution will be assigned a weight of 1, while weights will decrease for samples that are away from the main distribution.
The weight $w_s$ for each sample $s \in \mathcal{S}_i$ can be formulated as follows:

\begin{equation}
    w_{s}=\left\{\begin{array}{ll}
    1 & \text { if } d\left(s, \mathcal{D}_{0}\right) \leq \beta \\
    \exp \left(-\frac{d\left(s, \mathcal{D}_{0}\right)-\beta}{\sigma^{2}}\right) & \text { if } d\left(s, \mathcal{D}_{0}\right)>\beta
    \end{array}\right.
\end{equation}
where $d\left(s, \mathcal{D}_{0}\right)$ is a distance metric quantifying how far the sample $s$ is from the main distribution, $\beta$ is a predefined threshold distance, beyond which the weight begins to decrease,
$\sigma$ is a hyperparameter that controls the rate of decay for the weights of out-of-distribution samples.

The parameter update function can be written as follows:

\begin{equation}
    f(\theta, \mathcal{S}, \mathcal{P}) = \min_\theta\frac{1}{|\mathcal{S}|}\sum_{x_0 \in \mathcal{S}, c \in \mathcal{P}} \mathbb{E}\left [w_{x_0} \lambda_{t}||\hat{x_{\theta}}(x_{t}, t, c)- x_0) ||^ {2} \right ]
\end{equation}

With this weighting strategy, the in-distribution samples contribute optimally to model training while the others are penalized.
After a round of fine-tuning, we can evaluate the model and start a new training round. The overall framework is illustrated in \cref{fg:framework} and depicted in \cref{alg:rsidiff_procedure}.
\begin{figure}[t]
    \centering
    \includegraphics[width=0.26\linewidth]{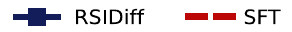}
    \subfloat[HPS test dataset.]{\includegraphics[width=0.48\linewidth]{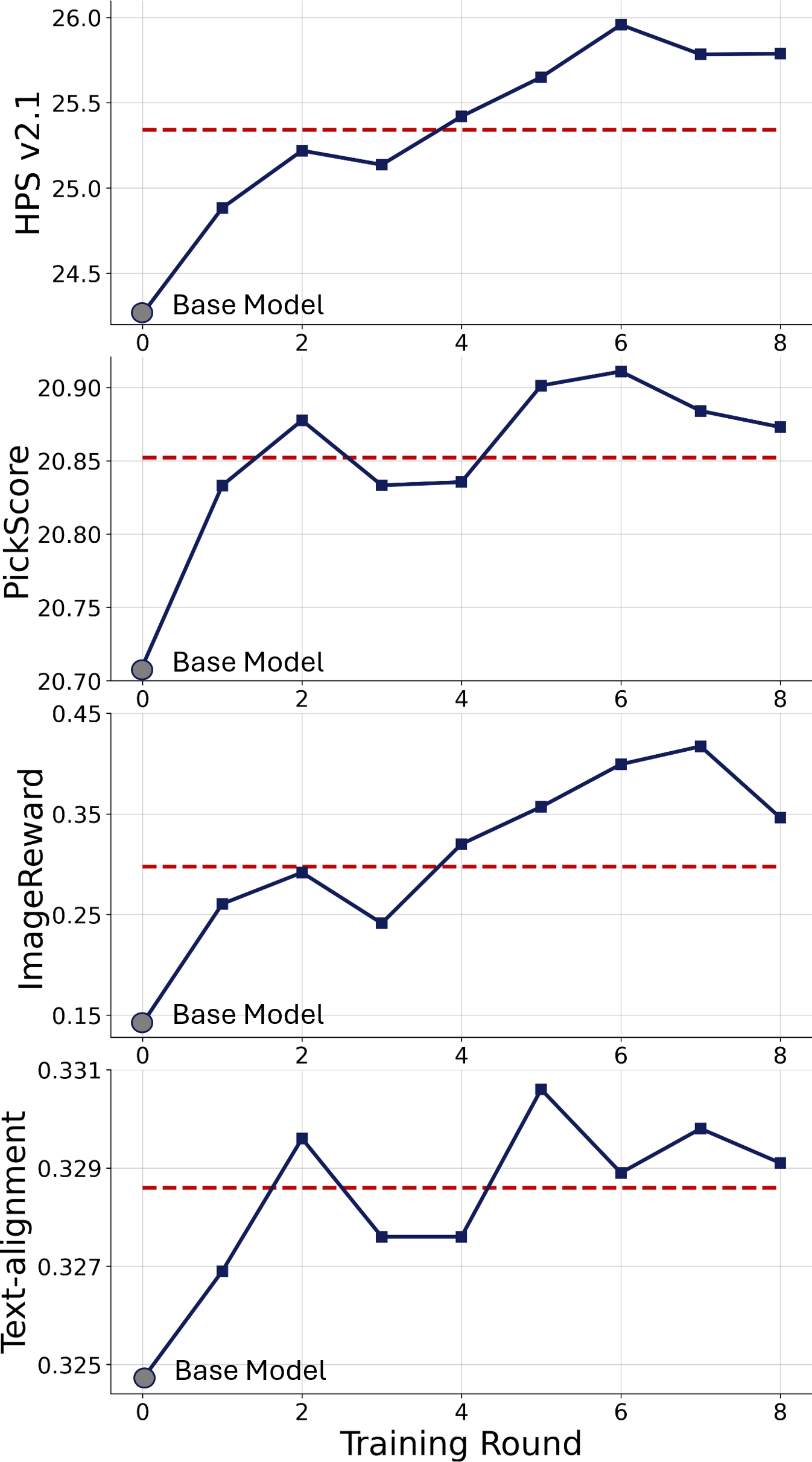}}
    \subfloat[PartiPrompts dataset.]{\includegraphics[width=0.48\linewidth]{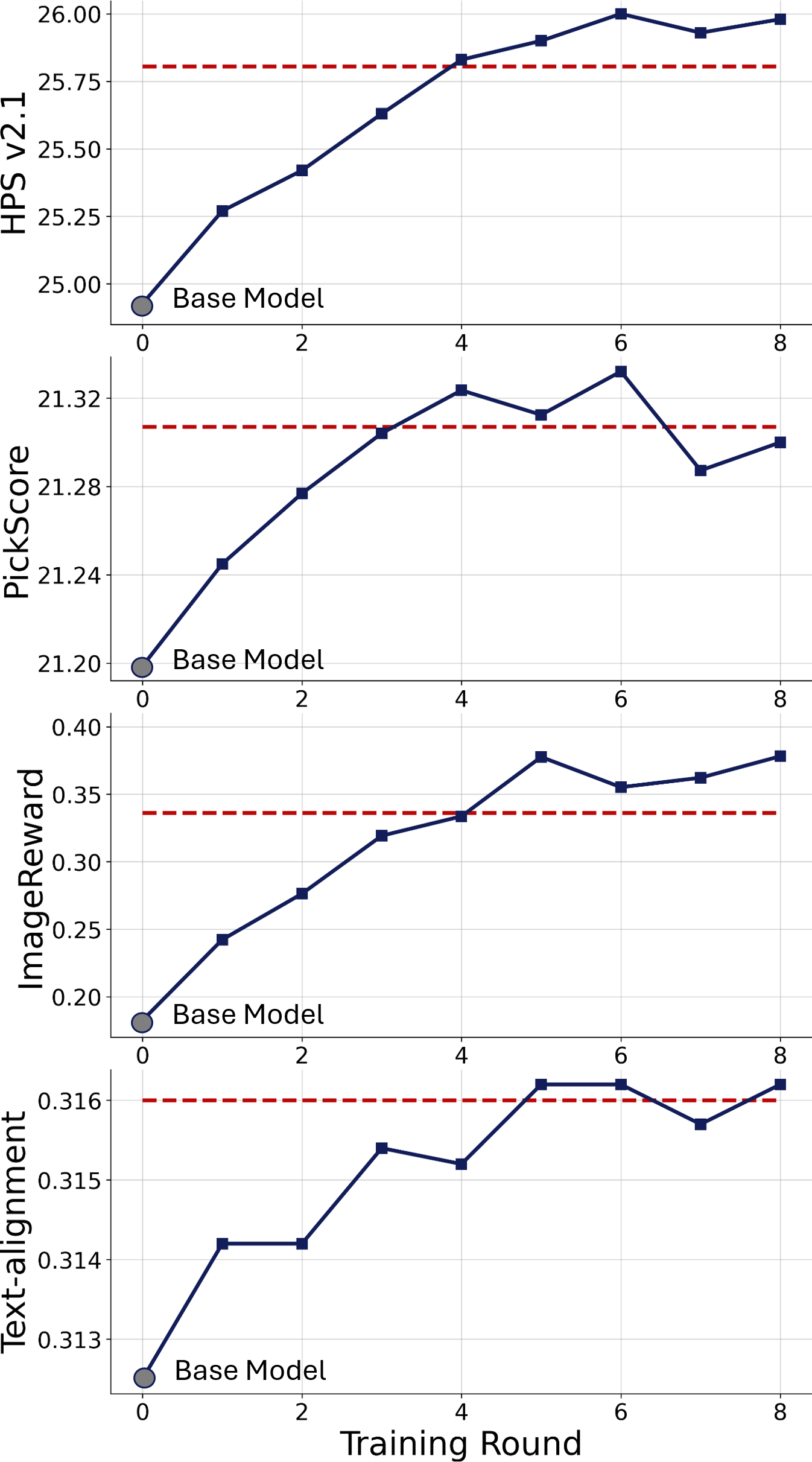}}
    \caption{\textbf{Quantitative Results.} Performance comparison with the base model and SFT method across two datasets and 4 evaluation metrics. The results show that RSIDiff significantly outperforms the base model and achieves consistent improvements.}
    \label{fg:precision}
\end{figure}


\begin{table}[t]
  \centering
  \begin{small}
  \begin{tabular}{lccc}
    \toprule
    Compared Method  & Visual Appeal & Text Faithfulness \\
    \midrule
    Base Model & 69.0\% & 77.2\% \\
    SFT & 59.8\% & 64.9\% \\
    \bottomrule
  \end{tabular}
  \caption{\textbf{User Study.} The percentage of user preference on RSIDiff (6th round) compared to the base model and SFT method.}
  \label{tb:user_study}
  \end{small}
\end{table}

\begin{figure*}[t]
\begin{center}
\centerline{\includegraphics[width=0.94\linewidth]{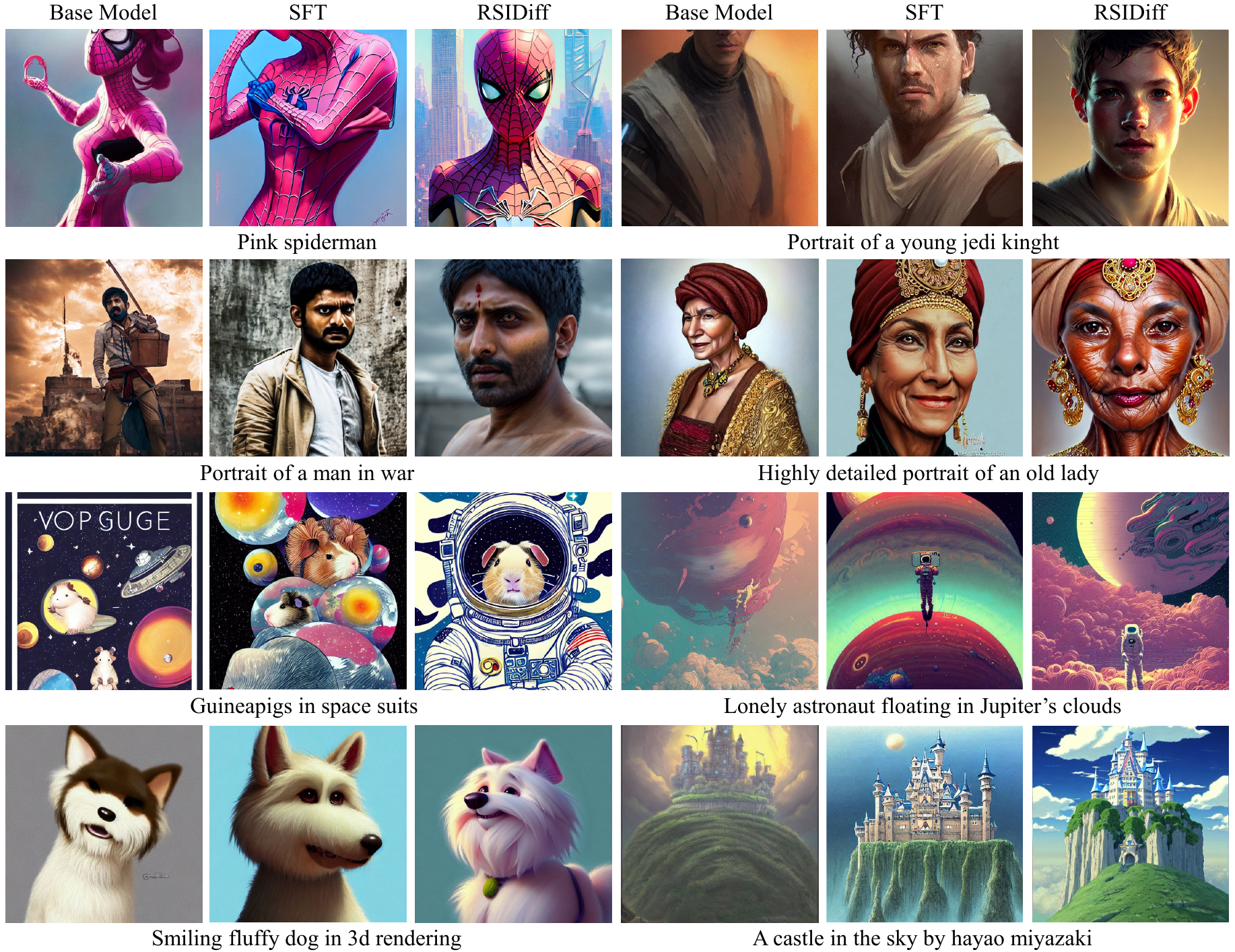}}
\caption{\textbf{Qualitative Results.} Examples generated by the base model (SD v1.4), SFT method, and RSIDiff at 6th round. The results illustrate RSIDiff's superior performance in several areas: effective centering of concepts (1st row), generation of intricate details (2nd row), better alignment with text prompts (3rd row), and enhanced understanding of artistic styles (4th row).}
\label{fg:sd_samples}
\end{center}
\vskip -0.2in
\end{figure*}

\vskip -0.1in
\section{Experiments}
\subsection{Experimental Setting}
\textbf{Datasets.} 
We conduct experiments using two representative datasets: PartiPrompts \cite{yu2022scaling} and HPS test \cite{wu2023human}.
The PartiPrompts dataset consists of 1,633 prompts that cover a variety of categories and challenges.
The HPS test dataset includes 3,200 prompts designed to evaluate four key aspects of text-to-image models: Animation, Concept Art, Painting, and Photo. For each prompt, we generate 10 images, resulting in a total of 4,8330 images per round.

\textbf{Evaluation Metrics.} We utilize four metrics: 
1) \textit{Text-alignment} evaluates how well the generated images correspond to textual descriptions. We implement it by measuring the similarity between the CLIP image features and the corresponding text features.
2) \textit{HPS v2} \cite{wu2023human} is an upgraded version of the human preference scorer \cite{wu2023humanhps}. In this paper, we utilize version 2.1.
3) \textit{PickScore} \cite{kirstain2023pick} is a CLIP-based preference scorer trained on the Pic-a-pic dataset \cite{kirstain2023pick}.
4) \textit{ImageReward} \cite{xu2024imagereward} is another preference scoring function.

\textbf{Hyperparameters.}
We use Stable Diffusion 1.4 \cite{rombach2022high} as the base model for self-improvement. In each training round, we generate images with 5k prompts and select 300 samples as the training set. The model is fine-tuned with batch size of 12 and learning rate of $1e^{-6}$ over 30 epochs. This process is recursively applied for a total of 8 rounds.

For the distribution-based weighting scheme, the parameters $\beta$ and $\sigma^2$ are set to 35 and 2, respectively. We employ the Variational Autoencoder (VAE) from SD v1.4 as the image encoder to calculate the mean value of distances between the selected samples and the reference dataset $\mathcal{D}_0$.

\begin{figure*}[t]
    \centering
    \includegraphics[width=0.7\linewidth]{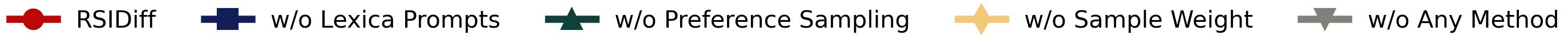}
    \subfloat[Evaluation on HPS test dataset.]{\includegraphics[width=0.93\linewidth]{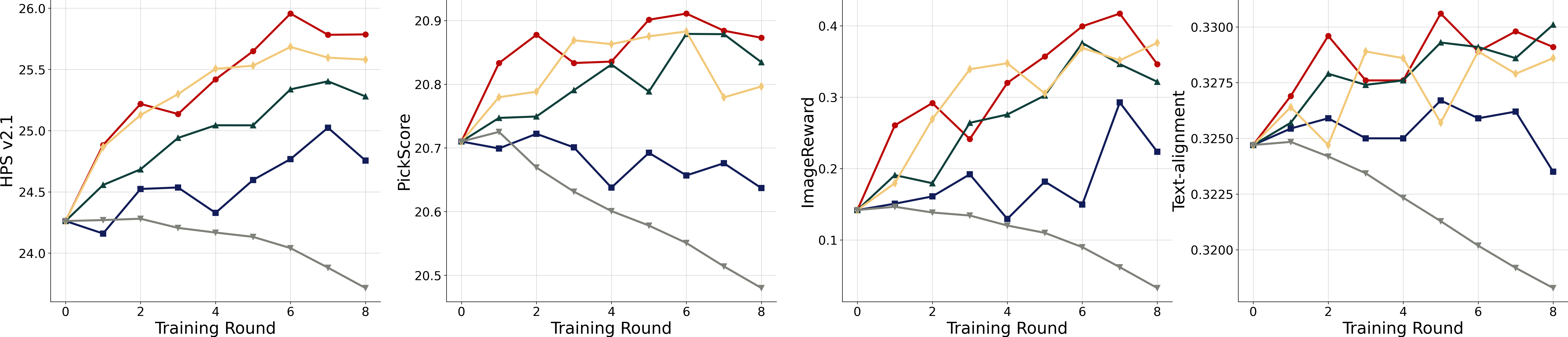}} \\
    \subfloat[Evaluation on PartiPrompts dataset.]{\includegraphics[width=0.93\linewidth]{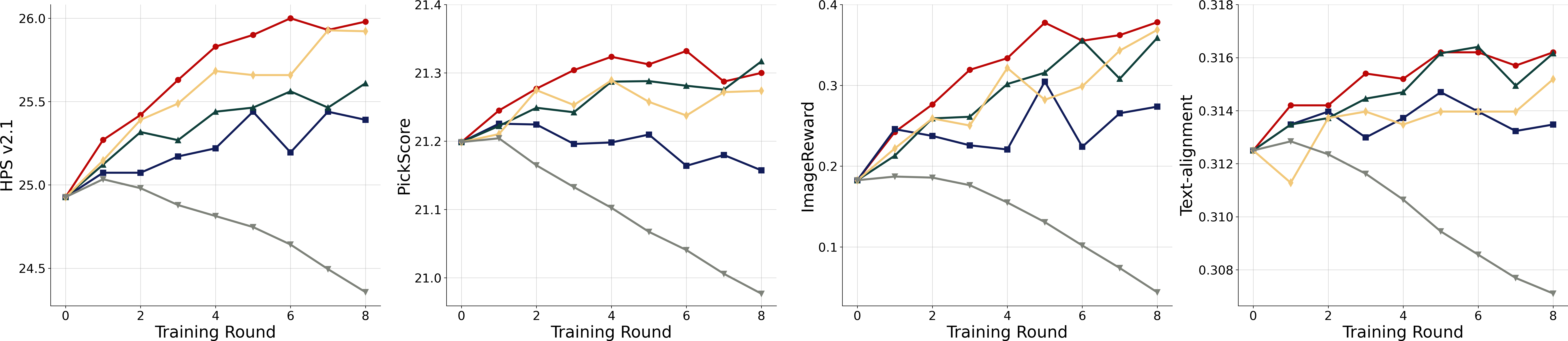}}
    \caption{\textbf{Ablation studies.} We compare RSIDiff with four configurations on the HPS test and PartiPrompts datasets. The results demonstrate that RSIDiff achieves superior recursive improvement while removing any individual strategy significantly degrades performance.}
    \label{fg:ablation}
\end{figure*}

\begin{figure}[t]
    \centering
    \includegraphics[width=0.7\linewidth]{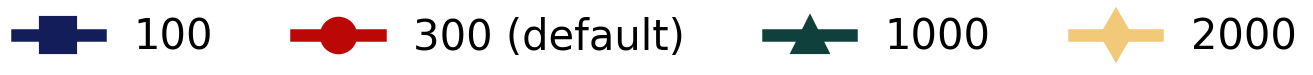}
    \includegraphics[width=0.96\linewidth]{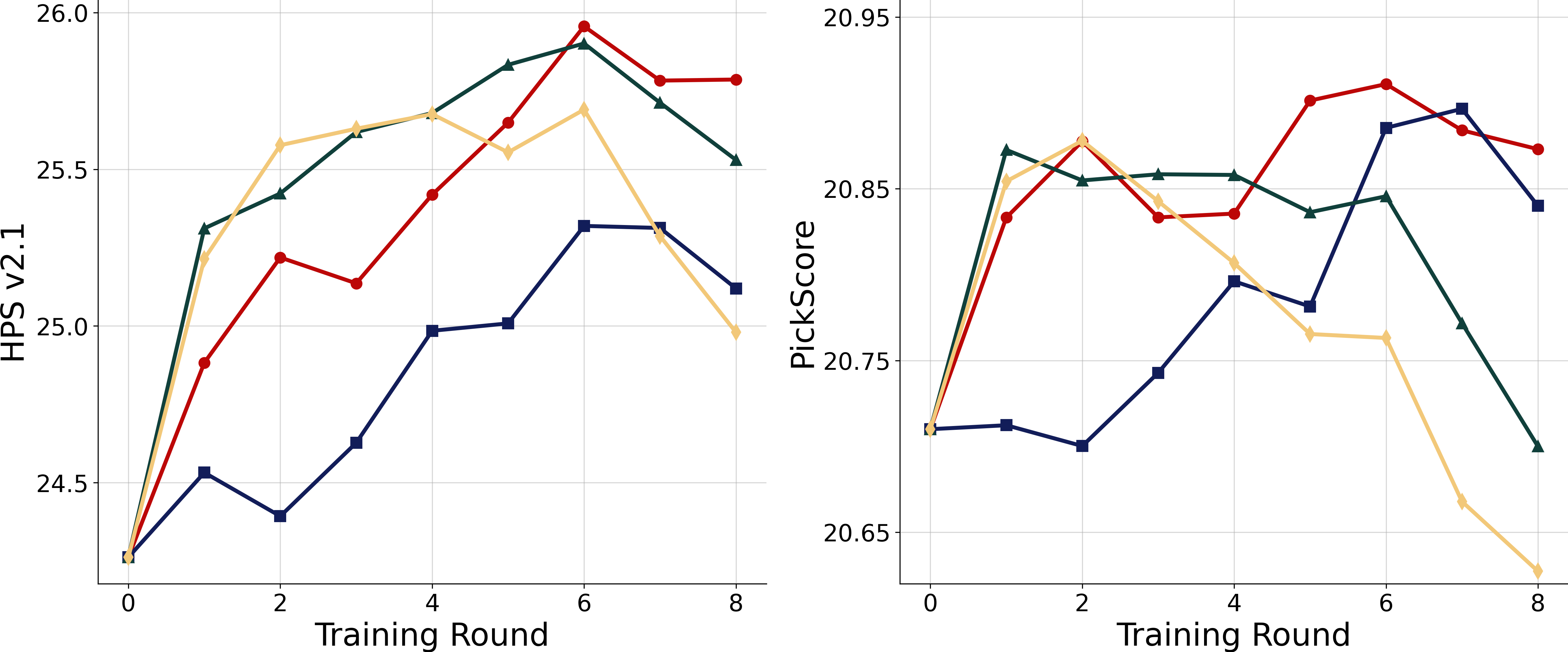}
    \caption{\textbf{Impact of sample size.} We select 100, 300, 1000, and 2000 samples from 5,000 synthetic data per training round and test the impact of sample size on the HPS test dataset.}
    \label{fg:sample_num}
    \vskip -0.1in
\end{figure}

\subsection{Results}


\textbf{Quantitative Results.}
We first validate the effectiveness of our proposed method by comparing it against the base model and supervised fine-tuning (SFT) method. 
For the SFT method, we fine-tune the diffusion model for 500 epochs using 5,000 synthetic data samples. Note that these data are generated by our filtered prompts, which have largely enhanced the aesthetic quality compared to poorly designed prompts. 
The results are illustrated in \cref{fg:precision}, where a training round of 0 reflects the performance of the baseline model.
It is clear that our method outperforms the base model and SFT method on all evaluation tests.
This demonstrates the effectiveness of RSIDiff.
%
Notably, the model achieves its highest performance in the 6th training round, surpassing the base model by 7.0\% and 181.6\% on the HPS v2.1 and ImageReward metrics of the HPS test dataset. 
Additionally, RSIDiff demonstrates superior performance compared to the SFT method, with advantages becoming evident after the 4th training round. 
Remarkably, this improvement is achieved using only 1,200 samples, significantly fewer than the scale of the SFT training set.
%

However, it is worth noting that the performance gained is not infinite. After the 6th round, we begin to see a decline on performance, suggesting that errors within the generated data start to adversely affect the model’s performance. This highlights the need for ongoing refinement and error management to sustain the benefits of self-improvement.

\textbf{User Study.}
We conduct a user study comprising two comparison tasks to evaluate the performance of RSIDiff. Participants are presented with two images generated by the same prompt with RSIDiff and the compared method. They are asked to evaluate the samples based on two key perspectives: Visual Appeal and Text Faithfulness.
This evaluation process yielded a total of 4,800 responses from 8 participants. The results are summarized in \cref{tb:user_study}.
It is clear that RSIDiff significantly improves both metrics.
Particularly, it outperforms the base model in Visual Appeal by 69.0\% and shows a substantial improvement by 59.8\% over the SFT method.
%
%
These findings suggest that RSIDiff enhances the base model by maintaining fidelity to the text and improving the overall visual quality of the generated images.

\textbf{Qualitative Results.} The qualitative results demonstrate that RSIDiff outperforms the base model, as illustrated in \cref{fg:sd_samples}. We highlight the key areas of improvement:
1) \textit{Concept Centering:} The first row shows RSIDiff produces well-crafted images rather than wrongly cropped photographs. 
2) \textit{Detail Rendering:} Our model generates intricate details, such as subtle facial features, as shown in the second row.
3) \textit{Text Alignment:} The third row illustrates that RSIDiff achieves superior alignment with the text prompts, like the successful generation of space suits, and a lonely astronaut.
4) \textit{Stylistic Understanding:} The fourth row highlights our method’s improved ability to understand painting styles, resulting in more aesthetically pleasing paintings.
Overall, these qualitative results underscore the effectiveness of RSIDiff in producing human-preferred images.

\subsection{Ablation Studies}
To validate the effectiveness of our proposed methods, we conduct ablation studies on both the HPS test dataset and the PartiPrompts dataset. 
We assess the impact of prompts by creating another prompt set with CLIP-based templates, such as ``a photo of [concept].''
We incorporate noun classes from the ImageNet dataset \cite{deng2009imagenet} to fill the [concept].
As illustrated in \cref{fg:ablation}, we visualize the impact of the three proposed methods alongside a setting without any proposed strategy.
Our results reveal that RSIDiff achieves the best recursive improvement compared to the other four settings. 
And we observe a noticeable performance drop when removing individual methods from RSIDiff.
Moreover, the results show a significant decline compared to the base model when no strategies are applied.
This indicates the importance of our proposed methods in mitigating generative hallucinations and enhancing the preference.

We next examine how the number of selected samples per training round affects performance. We compare four configurations: 100, 300 (default), 1000, and 2000. \cref{fg:sample_num} shows the results on the HPS test dataset.
Obviously, the 300-sample configuration significantly outperforms the 100-sample option by incorporating a larger number of valuable samples into the training process. 
However, the performance trends are less consistent when selecting 1000 or 2000 samples.
While these configurations exhibit superior performance in the early rounds, their improvement slows after round 3, even with a decline in PickScore.
This indicates that an excess of loosely selected samples may exacerbate the accumulation of generative hallucinations, potentially leading to an earlier happen of model collapse.

\begin{figure}[t]
\begin{center}
\centerline{\includegraphics[width=0.99\linewidth]{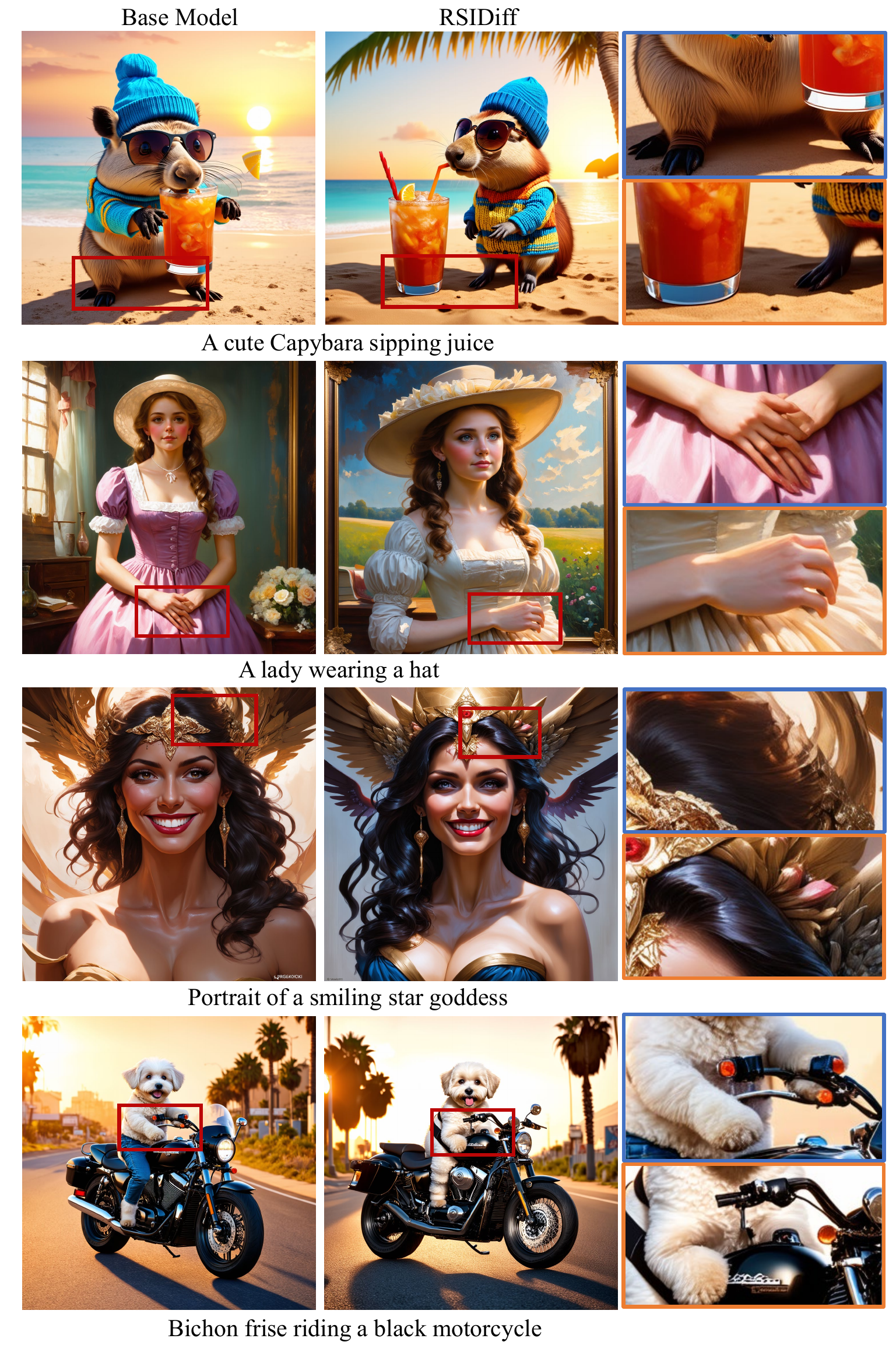}}
\caption{\textbf{Comparison with SD3.} We show the effectiveness of RSIDiff on SD3 in generating rational images (1st row), improving the detail in hand rendering (2nd row), depicting clear separation between hair and headdress (3rd row), and facilitating coherent interactions between the bichon and motorcycle (4th row).}
\label{fg:sd3_samples}
\end{center}
\vskip -0.2in
\end{figure}

\subsection{RSIDiff on SD3}
In this section, we evaluate the performance of RSIDiff by using the Stable Diffusion 3 medium (SD3) model \cite{esser2024scaling} as the base model.
During each preference sampling phase, we generate 13,000 images to create the synthetic data pool. After automated filtering with multiple metrics, we conduct an additional manual selection process to further filter the samples, ultimately obtaining 1,000 high-quality images. This ensures a more accurate alignment with human preferences. We optimize the LoRA \cite{hu2021lora} weights of SD3 with a learning rate of $1e^{-6}$ for 2,000 steps. We repeat this process for a total of 3 rounds.

Figure \ref{fg:sd3_samples} illustrates the comparison of generated images between the base model and RSIDiff. Our method corrects flaws present in the generated images. For example, RSIDiff demonstrates physically plausible generation, enhanced detail in hand rendering, clear separation between the character and the background, and more coherent interaction between subjects. These results validate our method for improving visual quality and adherence to user expectations.

\section{Conclusion}
This paper presents a study on the self-evolving text-to-image diffusion models. We highlight the degeneration issue when using self-generated data for self-training and attribute the problem to the lack of perceptual alignment and accumulation of generative hallucinations. To address this challenge, we introduce three key strategies: a prompt construction pipeline to enhance perceptual alignment, a preference sampling approach to prioritize human-aligned outputs while filtering out defective samples, and a distribution-based weighting mechanism to mitigate the effects of hallucinatory data. Our results show that RSIDiff significantly improves human preferences and model robustness. This work lays the groundwork for future research into enhancing diffusion models through RSI techniques.


{
    \small
    \bibliographystyle{ieeenat_fullname}
    \bibliography{cvpr}

\begin{thebibliography}{43}
\providecommand{\natexlab}[1]{#1}
\providecommand{\url}[1]{\texttt{#1}}
\expandafter\ifx\csname urlstyle\endcsname\relax
  \providecommand{\doi}[1]{doi: #1}\else
  \providecommand{\doi}{doi: \begingroup \urlstyle{rm}\Url}\fi

\bibitem[Ahmed et~al.(2020)Ahmed, Seraj, and Islam]{ahmed2020k}
Mohiuddin Ahmed, Raihan Seraj, and Syed Mohammed~Shamsul Islam.
\newblock The k-means algorithm: A comprehensive survey and performance evaluation.
\newblock \emph{Electronics}, 9\penalty0 (8):\penalty0 1295, 2020.

\bibitem[Alemohammad et~al.(2023)Alemohammad, Casco-Rodriguez, Luzi, Humayun, Babaei, LeJeune, Siahkoohi, and Baraniuk]{alemohammad2023self}
Sina Alemohammad, Josue Casco-Rodriguez, Lorenzo Luzi, Ahmed~Imtiaz Humayun, Hossein Babaei, Daniel LeJeune, Ali Siahkoohi, and Richard~G Baraniuk.
\newblock Self-consuming generative models go mad.
\newblock \emph{arXiv preprint arXiv:2307.01850}, 2023.

\bibitem[Betker et~al.(2023)Betker, Goh, Jing, Brooks, Wang, Li, Ouyang, Zhuang, Lee, Guo, et~al.]{betker2023improving}
James Betker, Gabriel Goh, Li Jing, Tim Brooks, Jianfeng Wang, Linjie Li, Long Ouyang, Juntang Zhuang, Joyce Lee, Yufei Guo, et~al.
\newblock Improving image generation with better captions.
\newblock \emph{Computer Science.}, 2\penalty0 (3):\penalty0 8, 2023.

\bibitem[Black et~al.(2023)Black, Janner, Du, Kostrikov, and Levine]{black2023training}
Kevin Black, Michael Janner, Yilun Du, Ilya Kostrikov, and Sergey Levine.
\newblock Training diffusion models with reinforcement learning.
\newblock \emph{arXiv preprint arXiv:2305.13301}, 2023.

\bibitem[Chen et~al.(2023)Chen, Lin, Sch{\"a}rli, and Zhou]{chen2023teaching}
Xinyun Chen, Maxwell Lin, Nathanael Sch{\"a}rli, and Denny Zhou.
\newblock Teaching large language models to self-debug.
\newblock \emph{arXiv preprint arXiv:2304.05128}, 2023.

\bibitem[Chen et~al.(2024)Chen, Deng, Yuan, Ji, and Gu]{chen2024self}
Zixiang Chen, Yihe Deng, Huizhuo Yuan, Kaixuan Ji, and Quanquan Gu.
\newblock Self-play fine-tuning converts weak language models to strong language models.
\newblock \emph{arXiv preprint arXiv:2401.01335}, 2024.

\bibitem[Dai et~al.(2023)Dai, Hou, Ma, Tsai, Wang, Wang, Zhang, Vandenhende, Wang, Dubey, et~al.]{dai2023emu}
Xiaoliang Dai, Ji Hou, Chih-Yao Ma, Sam Tsai, Jialiang Wang, Rui Wang, Peizhao Zhang, Simon Vandenhende, Xiaofang Wang, Abhimanyu Dubey, et~al.
\newblock Emu: Enhancing image generation models using photogenic needles in a haystack.
\newblock \emph{arXiv preprint arXiv:2309.15807}, 2023.

\bibitem[Deng et~al.(2009)Deng, Dong, Socher, Li, Li, and Fei-Fei]{deng2009imagenet}
Jia Deng, Wei Dong, Richard Socher, Li-Jia Li, Kai Li, and Li Fei-Fei.
\newblock Imagenet: A large-scale hierarchical image database.
\newblock In \emph{CVPR}, pages 248--255. Ieee, 2009.

\bibitem[Dohmatob et~al.(2024)Dohmatob, Feng, and Kempe]{dohmatob2024strong}
Elvis Dohmatob, Yunzhen Feng, and Julia Kempe.
\newblock Strong model collapse.
\newblock \emph{arXiv preprint arXiv:2410.04840}, 2024.

\bibitem[Dubey et~al.(2024)Dubey, Jauhri, Pandey, Kadian, Al-Dahle, Letman, Mathur, Schelten, Yang, Fan, et~al.]{dubey2024llama}
Abhimanyu Dubey, Abhinav Jauhri, Abhinav Pandey, Abhishek Kadian, Ahmad Al-Dahle, Aiesha Letman, Akhil Mathur, Alan Schelten, Amy Yang, Angela Fan, et~al.
\newblock The llama 3 herd of models.
\newblock \emph{arXiv preprint arXiv:2407.21783}, 2024.

\bibitem[Esser et~al.(2024)Esser, Kulal, Blattmann, Entezari, M{\"u}ller, Saini, Levi, Lorenz, Sauer, Boesel, et~al.]{esser2024scaling}
Patrick Esser, Sumith Kulal, Andreas Blattmann, Rahim Entezari, Jonas M{\"u}ller, Harry Saini, Yam Levi, Dominik Lorenz, Axel Sauer, Frederic Boesel, et~al.
\newblock Scaling rectified flow transformers for high-resolution image synthesis.
\newblock In \emph{ICML}, 2024.

\bibitem[Fan et~al.(2024)Fan, Watkins, Du, Liu, Ryu, Boutilier, Abbeel, Ghavamzadeh, Lee, and Lee]{fan2024reinforcement}
Ying Fan, Olivia Watkins, Yuqing Du, Hao Liu, Moonkyung Ryu, Craig Boutilier, Pieter Abbeel, Mohammad Ghavamzadeh, Kangwook Lee, and Kimin Lee.
\newblock Reinforcement learning for fine-tuning text-to-image diffusion models.
\newblock \emph{NIPS}, 36, 2024.

\bibitem[Gou et~al.(2023)Gou, Shao, Gong, Shen, Yang, Duan, and Chen]{gou2023critic}
Zhibin Gou, Zhihong Shao, Yeyun Gong, Yelong Shen, Yujiu Yang, Nan Duan, and Weizhu Chen.
\newblock Critic: Large language models can self-correct with tool-interactive critiquing.
\newblock \emph{arXiv preprint arXiv:2305.11738}, 2023.

\bibitem[Han et~al.(2024)Han, Liang, Shi, He, and Xiao]{han2024small}
Haixia Han, Jiaqing Liang, Jie Shi, Qianyu He, and Yanghua Xiao.
\newblock Small language model can self-correct.
\newblock In \emph{AAAI}, pages 18162--18170, 2024.

\bibitem[Ho et~al.(2020)Ho, Jain, and Abbeel]{ho2020denoising}
Jonathan Ho, Ajay Jain, and Pieter Abbeel.
\newblock Denoising diffusion probabilistic models.
\newblock \emph{NIPS}, 33:\penalty0 6840--6851, 2020.

\bibitem[Hu et~al.(2021)Hu, Shen, Wallis, Allen-Zhu, Li, Wang, Wang, and Chen]{hu2021lora}
Edward~J Hu, Yelong Shen, Phillip Wallis, Zeyuan Allen-Zhu, Yuanzhi Li, Shean Wang, Lu Wang, and Weizhu Chen.
\newblock Lora: Low-rank adaptation of large language models.
\newblock \emph{arXiv preprint arXiv:2106.09685}, 2021.

\bibitem[Karras et~al.(2020)Karras, Laine, Aittala, Hellsten, Lehtinen, and Aila]{karras2020analyzing}
Tero Karras, Samuli Laine, Miika Aittala, Janne Hellsten, Jaakko Lehtinen, and Timo Aila.
\newblock Analyzing and improving the image quality of stylegan.
\newblock In \emph{CVPR}, pages 8110--8119, 2020.

\bibitem[Kirstain et~al.(2023)Kirstain, Polyak, Singer, Matiana, Penna, and Levy]{kirstain2023pick}
Yuval Kirstain, Adam Polyak, Uriel Singer, Shahbuland Matiana, Joe Penna, and Omer Levy.
\newblock Pick-a-pic: An open dataset of user preferences for text-to-image generation.
\newblock \emph{NIPS}, 36:\penalty0 36652--36663, 2023.

\bibitem[Madaan et~al.(2024)Madaan, Tandon, Gupta, Hallinan, Gao, Wiegreffe, Alon, Dziri, Prabhumoye, Yang, et~al.]{madaan2024self}
Aman Madaan, Niket Tandon, Prakhar Gupta, Skyler Hallinan, Luyu Gao, Sarah Wiegreffe, Uri Alon, Nouha Dziri, Shrimai Prabhumoye, Yiming Yang, et~al.
\newblock Self-refine: Iterative refinement with self-feedback.
\newblock \emph{NIPS}, 36, 2024.

\bibitem[Miao et~al.(2023)Miao, Teh, and Rainforth]{miao2023selfcheck}
Ning Miao, Yee~Whye Teh, and Tom Rainforth.
\newblock Selfcheck: Using llms to zero-shot check their own step-by-step reasoning.
\newblock \emph{arXiv preprint arXiv:2308.00436}, 2023.

\bibitem[Nivel et~al.(2013)Nivel, Th{\'o}risson, Steunebrink, Dindo, Pezzulo, Rodriguez, Hern{\'a}ndez, Ognibene, Schmidhuber, Sanz, et~al.]{nivel2013bounded}
Eric Nivel, Kristinn~R Th{\'o}risson, Bas~R Steunebrink, Haris Dindo, Giovanni Pezzulo, Manuel Rodriguez, Carlos Hern{\'a}ndez, Dimitri Ognibene, J{\"u}rgen Schmidhuber, Ricardo Sanz, et~al.
\newblock Bounded recursive self-improvement.
\newblock \emph{arXiv preprint arXiv:1312.6764}, 2013.

\bibitem[Podell et~al.(2023)Podell, English, Lacey, Blattmann, Dockhorn, Müller, Penna, and Rombach]{podell2023sdxl}
Dustin Podell, Zion English, Kyle Lacey, Andreas Blattmann, Tim Dockhorn, Jonas Müller, Joe Penna, and Robin Rombach.
\newblock Sdxl: Improving latent diffusion models for high-resolution image synthesis.
\newblock \emph{arXiv preprint arXiv:2307.01952}, 2023.

\bibitem[Radford et~al.(2021)Radford, Kim, Hallacy, Ramesh, Goh, Agarwal, Sastry, Askell, Mishkin, Clark, et~al.]{radford2021learning}
Alec Radford, Jong~Wook Kim, Chris Hallacy, Aditya Ramesh, Gabriel Goh, Sandhini Agarwal, Girish Sastry, Amanda Askell, Pamela Mishkin, Jack Clark, et~al.
\newblock Learning transferable visual models from natural language supervision.
\newblock In \emph{ICML}, pages 8748--8763. PMLR, 2021.

\bibitem[Rafailov et~al.(2024)Rafailov, Sharma, Mitchell, Manning, Ermon, and Finn]{rafailov2024direct}
Rafael Rafailov, Archit Sharma, Eric Mitchell, Christopher~D Manning, Stefano Ermon, and Chelsea Finn.
\newblock Direct preference optimization: Your language model is secretly a reward model.
\newblock \emph{NIPS}, 36, 2024.

\bibitem[Ramesh et~al.(2022)Ramesh, Dhariwal, Nichol, Chu, and Chen]{ramesh2022hierarchical}
Aditya Ramesh, Prafulla Dhariwal, Alex Nichol, Casey Chu, and Mark Chen.
\newblock Hierarchical text-conditional image generation with clip latents.
\newblock \emph{arXiv preprint arXiv:2204.06125}, 1\penalty0 (2):\penalty0 3, 2022.

\bibitem[Rombach et~al.(2022)Rombach, Blattmann, Lorenz, Esser, and Ommer]{rombach2022high}
Robin Rombach, Andreas Blattmann, Dominik Lorenz, Patrick Esser, and Bj{\"o}rn Ommer.
\newblock High-resolution image synthesis with latent diffusion models.
\newblock In \emph{CVPR}, pages 10684--10695, 2022.

\bibitem[Schmidhuber(2003)]{schmidhuber2003godel}
J{\"u}rgen Schmidhuber.
\newblock G{\"o}del machines: self-referential universal problem solvers making provably optimal self-improvements.
\newblock \emph{arXiv preprint cs/0309048}, 2003.

\bibitem[Schuhmann et~al.(2022)Schuhmann, Beaumont, Vencu, Gordon, Wightman, Cherti, Coombes, Katta, Mullis, Wortsman, et~al.]{schuhmann2022laion}
Christoph Schuhmann, Romain Beaumont, Richard Vencu, Cade Gordon, Ross Wightman, Mehdi Cherti, Theo Coombes, Aarush Katta, Clayton Mullis, Mitchell Wortsman, et~al.
\newblock Laion-5b: An open large-scale dataset for training next generation image-text models.
\newblock \emph{NIPS}, 35:\penalty0 25278--25294, 2022.

\bibitem[Shumailov et~al.(2024)Shumailov, Shumaylov, Zhao, Papernot, Anderson, and Gal]{shumailov2024ai}
Ilia Shumailov, Zakhar Shumaylov, Yiren Zhao, Nicolas Papernot, Ross Anderson, and Yarin Gal.
\newblock Ai models collapse when trained on recursively generated data.
\newblock \emph{Nature}, 631\penalty0 (8022):\penalty0 755--759, 2024.

\bibitem[Silver et~al.(2016)Silver, Huang, Maddison, Guez, Sifre, Van Den~Driessche, Schrittwieser, Antonoglou, Panneershelvam, Lanctot, et~al.]{silver2016mastering}
David Silver, Aja Huang, Chris~J Maddison, Arthur Guez, Laurent Sifre, George Van Den~Driessche, Julian Schrittwieser, Ioannis Antonoglou, Veda Panneershelvam, Marc Lanctot, et~al.
\newblock Mastering the game of go with deep neural networks and tree search.
\newblock \emph{Nature}, 529\penalty0 (7587):\penalty0 484--489, 2016.

\bibitem[Steunebrink et~al.(2016)Steunebrink, Th{\'o}risson, and Schmidhuber]{steunebrink2016growing}
Bas~R Steunebrink, Kristinn~R Th{\'o}risson, and J{\"u}rgen Schmidhuber.
\newblock Growing recursive self-improvers.
\newblock In \emph{ICAGI}, pages 129--139. Springer, 2016.

\bibitem[Tao et~al.(2024)Tao, Lin, Chen, Li, Wu, Li, Jin, Huang, Tao, and Zhou]{tao2024survey}
Zhengwei Tao, Ting-En Lin, Xiancai Chen, Hangyu Li, Yuchuan Wu, Yongbin Li, Zhi Jin, Fei Huang, Dacheng Tao, and Jingren Zhou.
\newblock A survey on self-evolution of large language models.
\newblock \emph{arXiv preprint arXiv:2404.14387}, 2024.

\bibitem[Wallace et~al.(2024)Wallace, Dang, Rafailov, Zhou, Lou, Purushwalkam, Ermon, Xiong, Joty, and Naik]{wallace2024diffusion}
Bram Wallace, Meihua Dang, Rafael Rafailov, Linqi Zhou, Aaron Lou, Senthil Purushwalkam, Stefano Ermon, Caiming Xiong, Shafiq Joty, and Nikhil Naik.
\newblock Diffusion model alignment using direct preference optimization.
\newblock In \emph{CVPR}, pages 8228--8238, 2024.

\bibitem[Welleck et~al.()Welleck, Lu, West, Brahman, Shen, Khashabi, and Choi]{welleckgenerating}
Sean Welleck, Ximing Lu, Peter West, Faeze Brahman, Tianxiao Shen, Daniel Khashabi, and Yejin Choi.
\newblock Generating sequences by learning to self-correct.
\newblock In \emph{ICLR}.

\bibitem[Wu et~al.(2023{\natexlab{a}})Wu, Hao, Sun, Chen, Zhu, Zhao, and Li]{wu2023human}
Xiaoshi Wu, Yiming Hao, Keqiang Sun, Yixiong Chen, Feng Zhu, Rui Zhao, and Hongsheng Li.
\newblock Human preference score v2: A solid benchmark for evaluating human preferences of text-to-image synthesis.
\newblock \emph{arXiv preprint arXiv:2306.09341}, 2023{\natexlab{a}}.

\bibitem[Wu et~al.(2023{\natexlab{b}})Wu, Sun, Zhu, Zhao, and Li]{wu2023humanhps}
Xiaoshi Wu, Keqiang Sun, Feng Zhu, Rui Zhao, and Hongsheng Li.
\newblock Human preference score: Better aligning text-to-image models with human preference.
\newblock In \emph{ICCV}, pages 2096--2105, 2023{\natexlab{b}}.

\bibitem[Xu et~al.(2024)Xu, Liu, Wu, Tong, Li, Ding, Tang, and Dong]{xu2024imagereward}
Jiazheng Xu, Xiao Liu, Yuchen Wu, Yuxuan Tong, Qinkai Li, Ming Ding, Jie Tang, and Yuxiao Dong.
\newblock Imagereward: Learning and evaluating human preferences for text-to-image generation.
\newblock \emph{NIPS}, 36, 2024.

\bibitem[Yampolskiy(2015)]{yampolskiy2015seed}
Roman~V Yampolskiy.
\newblock From seed ai to technological singularity via recursively self-improving software.
\newblock \emph{arXiv preprint arXiv:1502.06512}, 2015.

\bibitem[Yang et~al.(2024)Yang, Tao, Lyu, Ge, Chen, Shen, Zhu, and Li]{yang2024using}
Kai Yang, Jian Tao, Jiafei Lyu, Chunjiang Ge, Jiaxin Chen, Weihan Shen, Xiaolong Zhu, and Xiu Li.
\newblock Using human feedback to fine-tune diffusion models without any reward model.
\newblock In \emph{CVPR}, pages 8941--8951, 2024.

\bibitem[Yu et~al.(2022)Yu, Xu, Koh, Luong, Baid, Wang, Vasudevan, Ku, Yang, Ayan, et~al.]{yu2022scaling}
Jiahui Yu, Yuanzhong Xu, Jing~Yu Koh, Thang Luong, Gunjan Baid, Zirui Wang, Vijay Vasudevan, Alexander Ku, Yinfei Yang, Burcu~Karagol Ayan, et~al.
\newblock Scaling autoregressive models for content-rich text-to-image generation.
\newblock \emph{arXiv preprint arXiv:2206.10789}, 2\penalty0 (3):\penalty0 5, 2022.

\bibitem[Yuan et~al.(2024)Yuan, Pang, Cho, Sukhbaatar, Xu, and Weston]{yuan2024self}
Weizhe Yuan, Richard~Yuanzhe Pang, Kyunghyun Cho, Sainbayar Sukhbaatar, Jing Xu, and Jason Weston.
\newblock Self-rewarding language models.
\newblock \emph{arXiv preprint arXiv:2401.10020}, 2024.

\bibitem[Zelikman et~al.(2022)Zelikman, Wu, Mu, and Goodman]{zelikman2022star}
Eric Zelikman, Yuhuai Wu, Jesse Mu, and Noah Goodman.
\newblock Star: Bootstrapping reasoning with reasoning.
\newblock \emph{NIPS}, 35:\penalty0 15476--15488, 2022.

\bibitem[Zhang et~al.(2022)Zhang, Roller, Goyal, Artetxe, Chen, Chen, Dewan, Diab, Li, Lin, et~al.]{zhang2022opt}
Susan Zhang, Stephen Roller, Naman Goyal, Mikel Artetxe, Moya Chen, Shuohui Chen, Christopher Dewan, Mona Diab, Xian Li, Xi~Victoria Lin, et~al.
\newblock Opt: Open pre-trained transformer language models.
\newblock \emph{arXiv preprint arXiv:2205.01068}, 2022.

\end{thebibliography}
}



\section{Supplementary Materials}
To provide a more comprehensive understanding of the method, we have included additional details in the following sections. The source code can be accessed at https://open\_upon\_acceptance.

\subsection{Prompt Set Examples}
This section presents partial prompt samples for generating human-aligned images, as shown in \cref{tb:prompt}. Our prompt set demonstrates key qualities: clarity, specificity, and diversity.

\subsection{More Ablation Studies}
\begin{figure*}[t]
    \centering
    \includegraphics[width=0.4\linewidth]{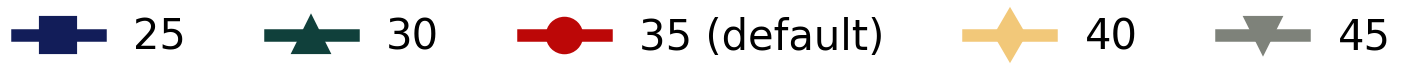}
    \subfloat[Evaluation on HPS test dataset.]{\includegraphics[width=0.93\linewidth]{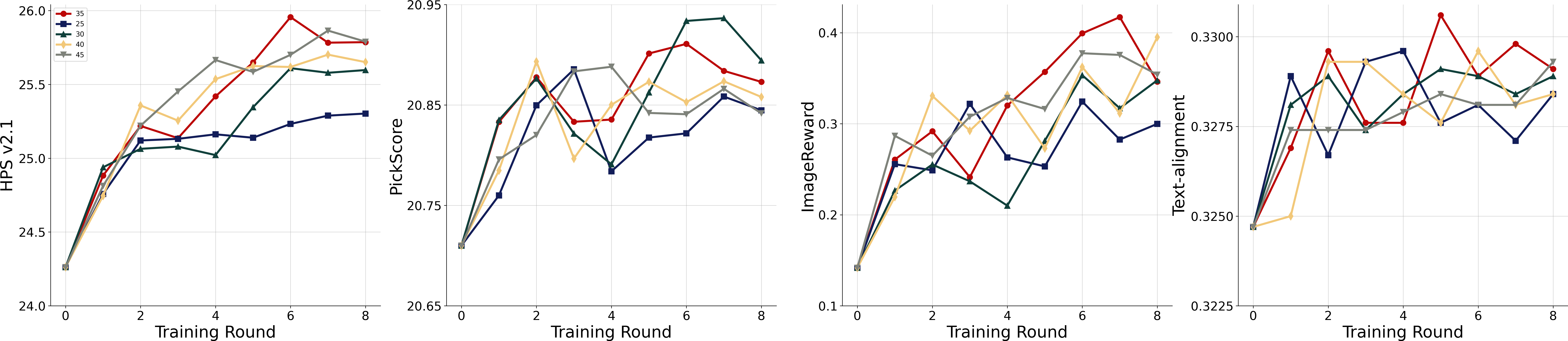}} \\
    \subfloat[Evaluation on PartiPrompts dataset.]{\includegraphics[width=0.93\linewidth]{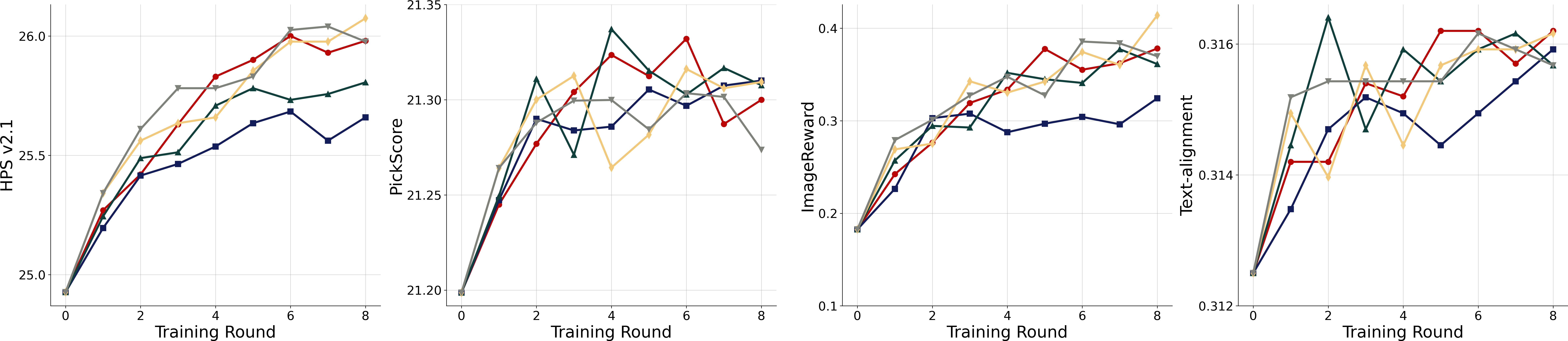}}
    \caption{\textbf{Ablation Study of $\beta$.} We assess the effects of $\beta$ under four different metrics on the HPS test and PartiPrompts datasets. Higher values of $\beta$ correspond to a more permissive determination of in-distribution samples. We set $\beta$ as 35 based on the overall performance.}
    \label{fg:beta_ablation}
\end{figure*}

\begin{figure*}[t]
    \centering
    \includegraphics[width=0.4\linewidth]{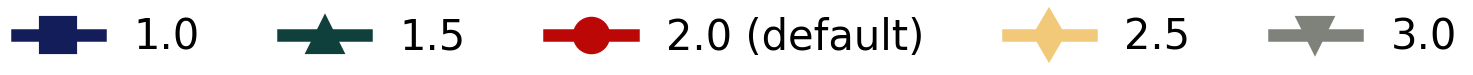}
    \subfloat[Evaluation on HPS test dataset.]{\includegraphics[width=0.93\linewidth]{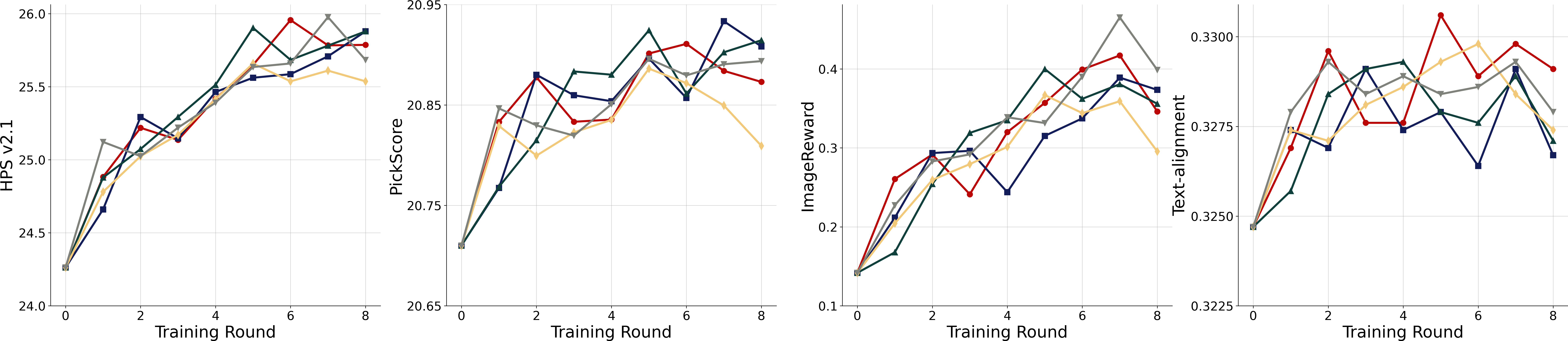}} \\
    \subfloat[Evaluation on PartiPrompts dataset.]{\includegraphics[width=0.93\linewidth]{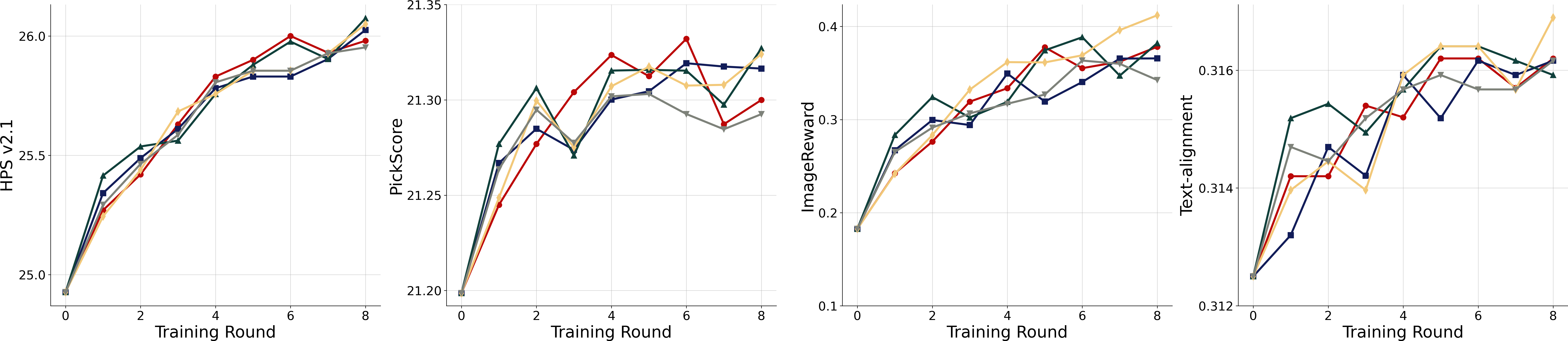}}
    \caption{\textbf{Ablation Study of $\sigma^2$.} We assess the effects of $\sigma^2$ under four different metrics on the HPS test and PartiPrompts datasets. Smaller values of $\sigma^2$ correspond to a higher penalization on out-of-distribution samples. We set $\sigma^2$ as 2.0 based on the overall performance.}
    \label{fg:sigma_ablation}
\end{figure*}

In this section, we conduct an ablation study to analyze the impact of the hyperparameters $\beta$ and $\sigma$ as defined in Eq. (4). The parameter $\beta$ determines whether samples are in-distribution or out-of-distribution, while $\sigma$ influences the penalization weight assigned to out-of-distribution samples. Results are shown in \cref{fg:beta_ablation} and \cref{fg:sigma_ablation}. It is clear that model performance varies on these parameters, which indicates the importance of the distribution-based weighting scheme. Based on overall results, we set $\beta$ and $\sigma^2$ to 35 and 2, respectively.

\subsection{Additional Results}
\begin{figure*}[t]
\begin{center}
\centerline{\includegraphics[width=0.98\linewidth]{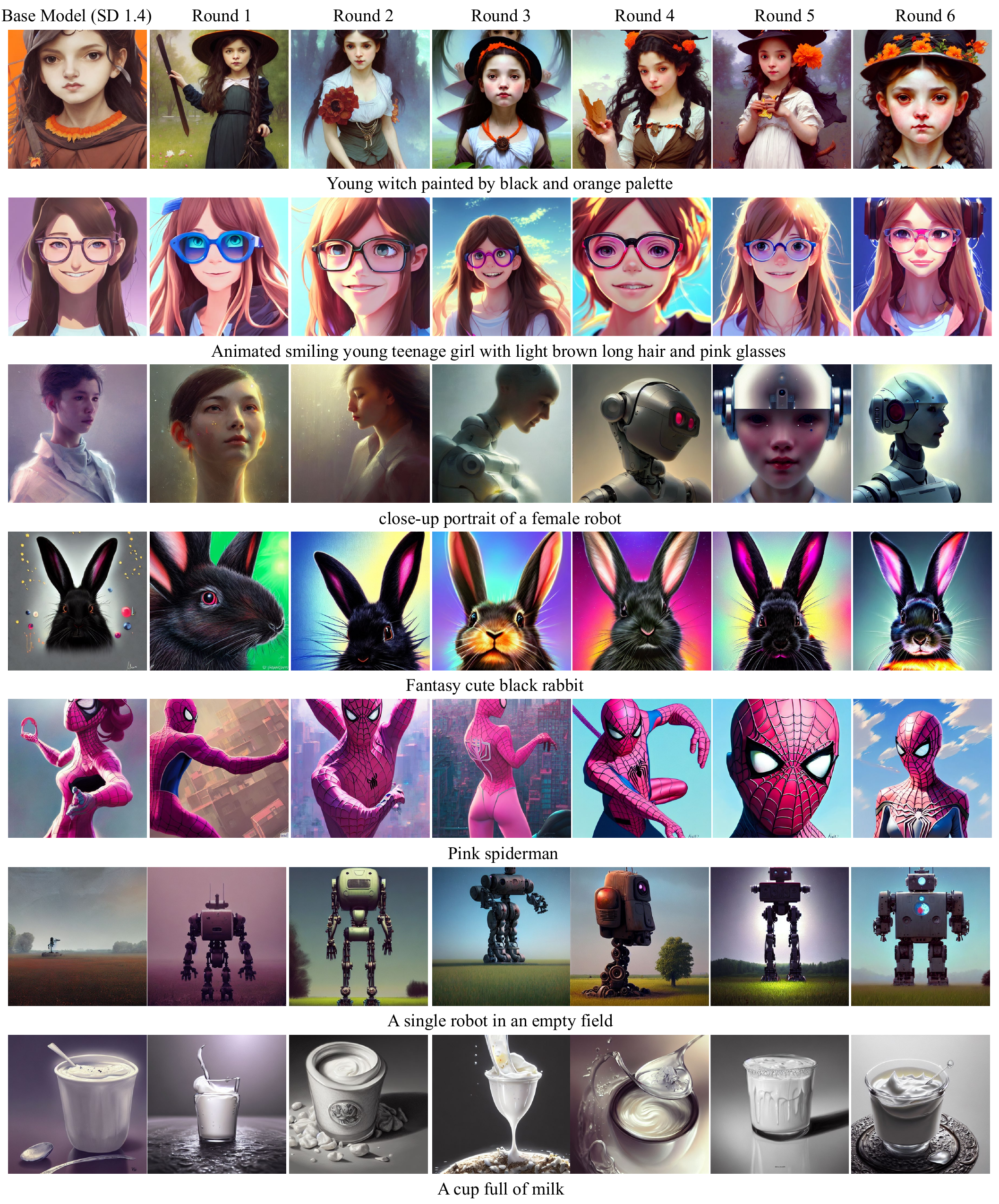}}
\caption{\textbf{Examples Generated in Rounds.} We show examples generated from rounds 1 to 6, alongside outputs from the base model (SD v1.4). The images illustrate a gradual enhancement in visual quality.}
\label{fg:supp_iter}
\end{center}
\end{figure*}

\begin{figure*}[t]
\begin{center}
\centerline{\includegraphics[width=0.9\linewidth]{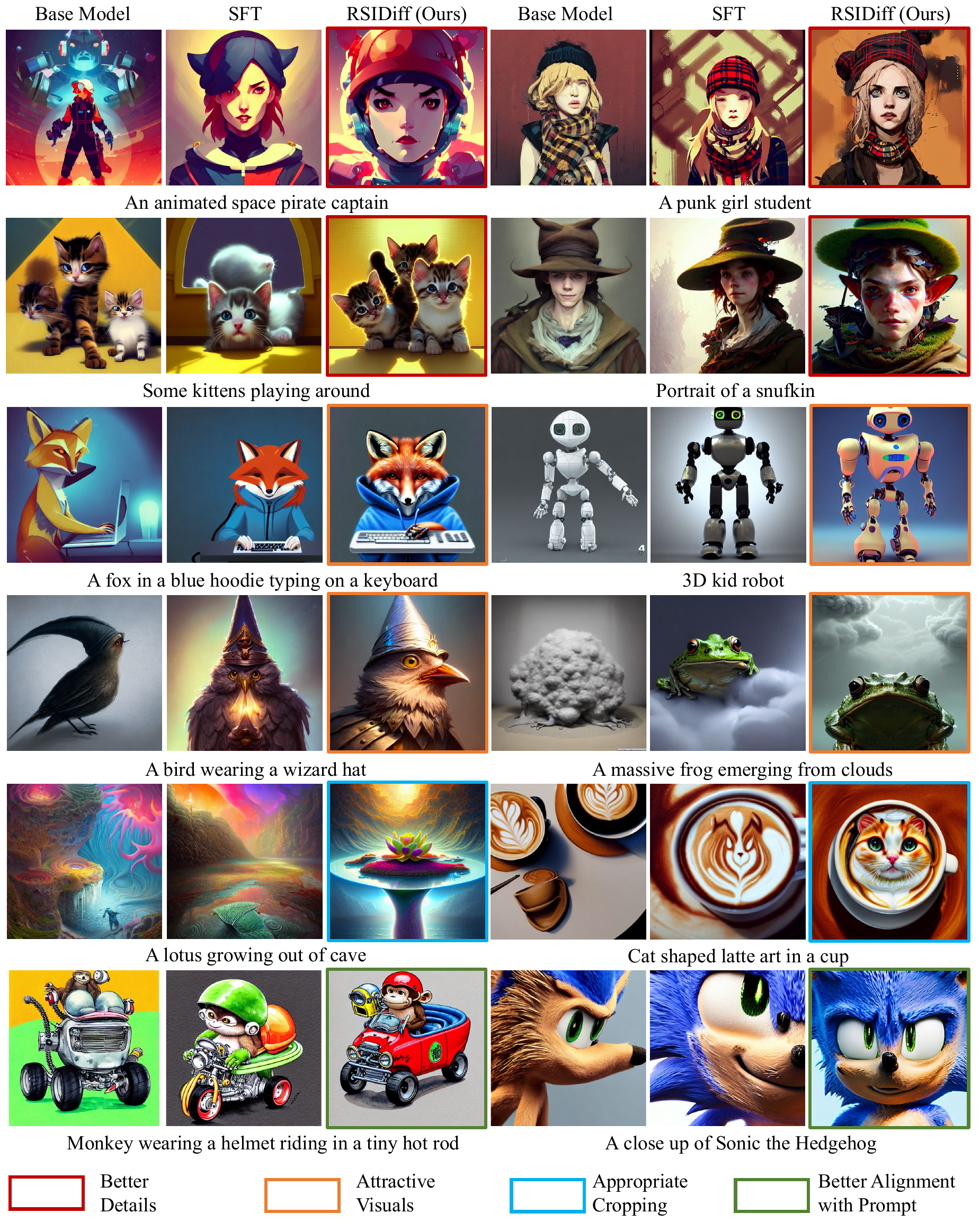}}
\caption{\textbf{Qualitative Results.} Examples generated by the base model (SD v1.4), SFT method, and RSIDiff. The results illustrate RSIDiff's superior performance in several areas: generation of intricate details, effective centering of concepts, human-aligned aesthetic understanding, and better alignment with text prompts.}
\label{fg:supp_sd_samples}
\end{center}
\end{figure*}

\begin{figure*}[t]
\begin{center}
\centerline{\includegraphics[width=0.75\linewidth]{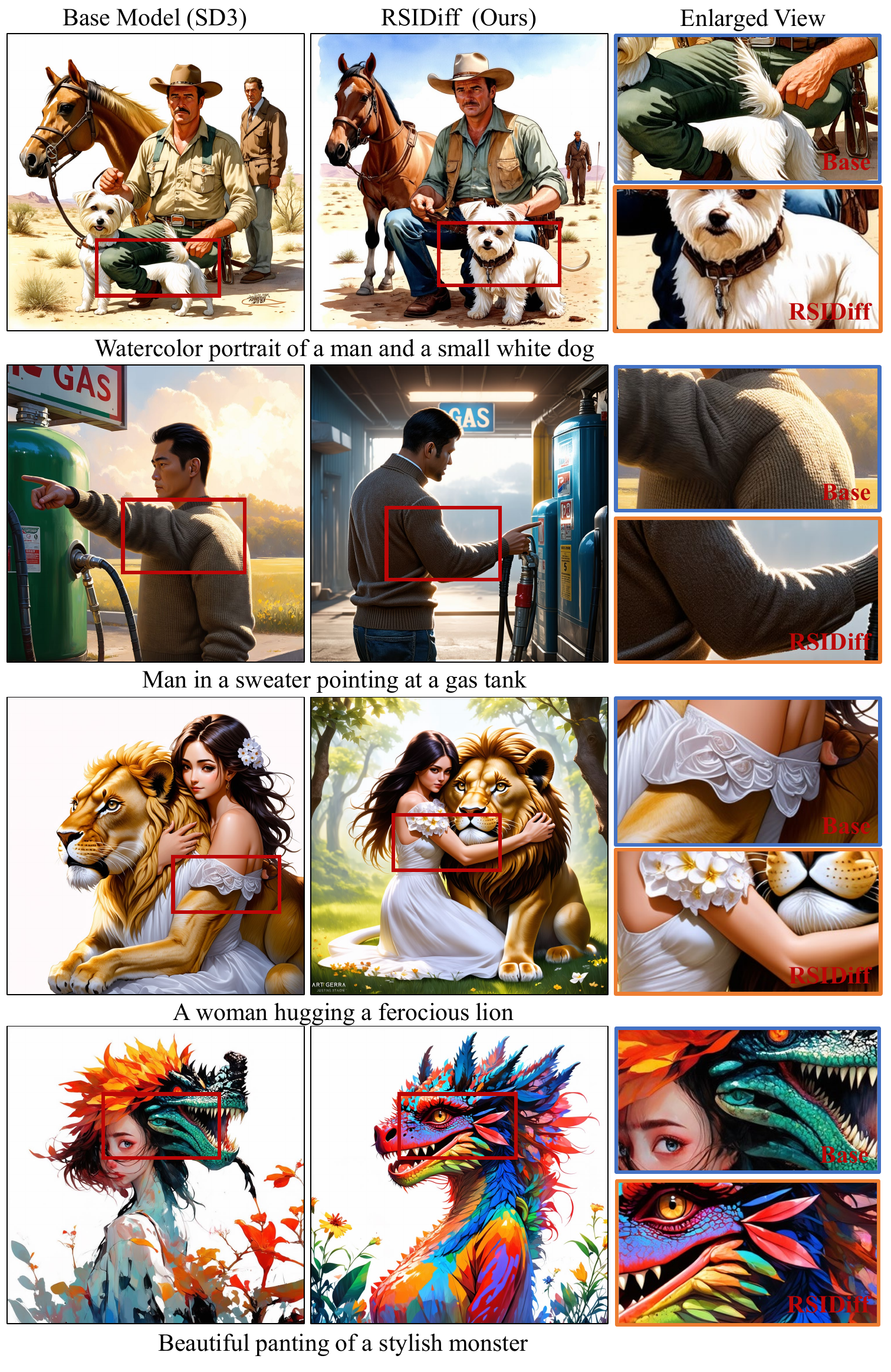}}
\caption{\textbf{Examples Generated by SD3 and RSIDiff.} This comparison highlights that our method RSIDiff enhances the base model by achieving more coherent interactions between subjects.}
\label{fg:supp_sd3_1}
\end{center}
\end{figure*}

\begin{figure*}[t]
\begin{center}
\centerline{\includegraphics[width=0.75\linewidth]{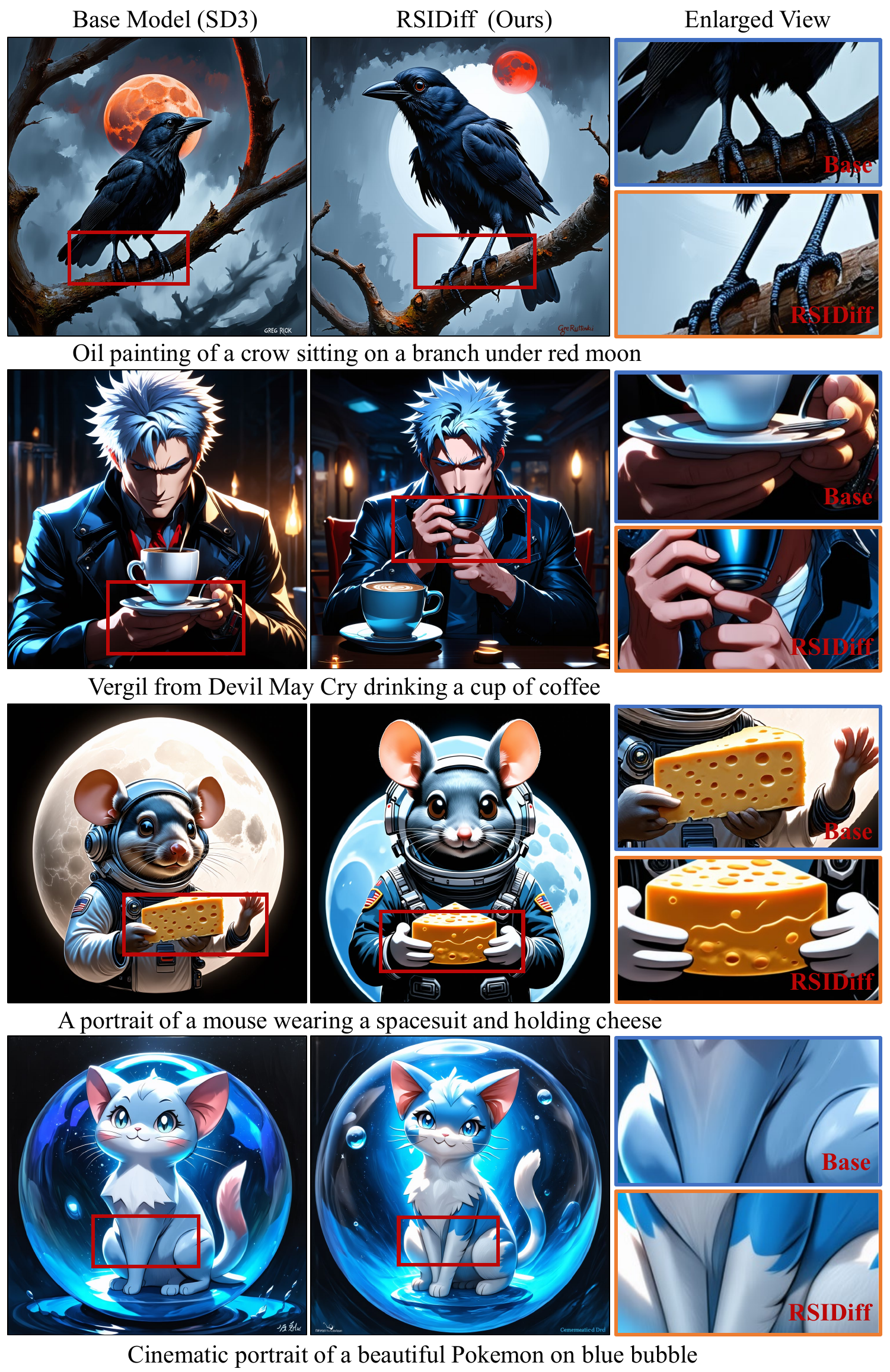}}
\caption{\textbf{Examples Generated by SD3 and RSIDiff.} This comparison highlights that our method RSIDiff enhances the base model by improving
the detail in hands and feet rendering.}
\label{fg:supp_sd3_2}
\end{center}
\end{figure*}

\begin{figure*}[t]
\begin{center}
\centerline{\includegraphics[width=0.75\linewidth]{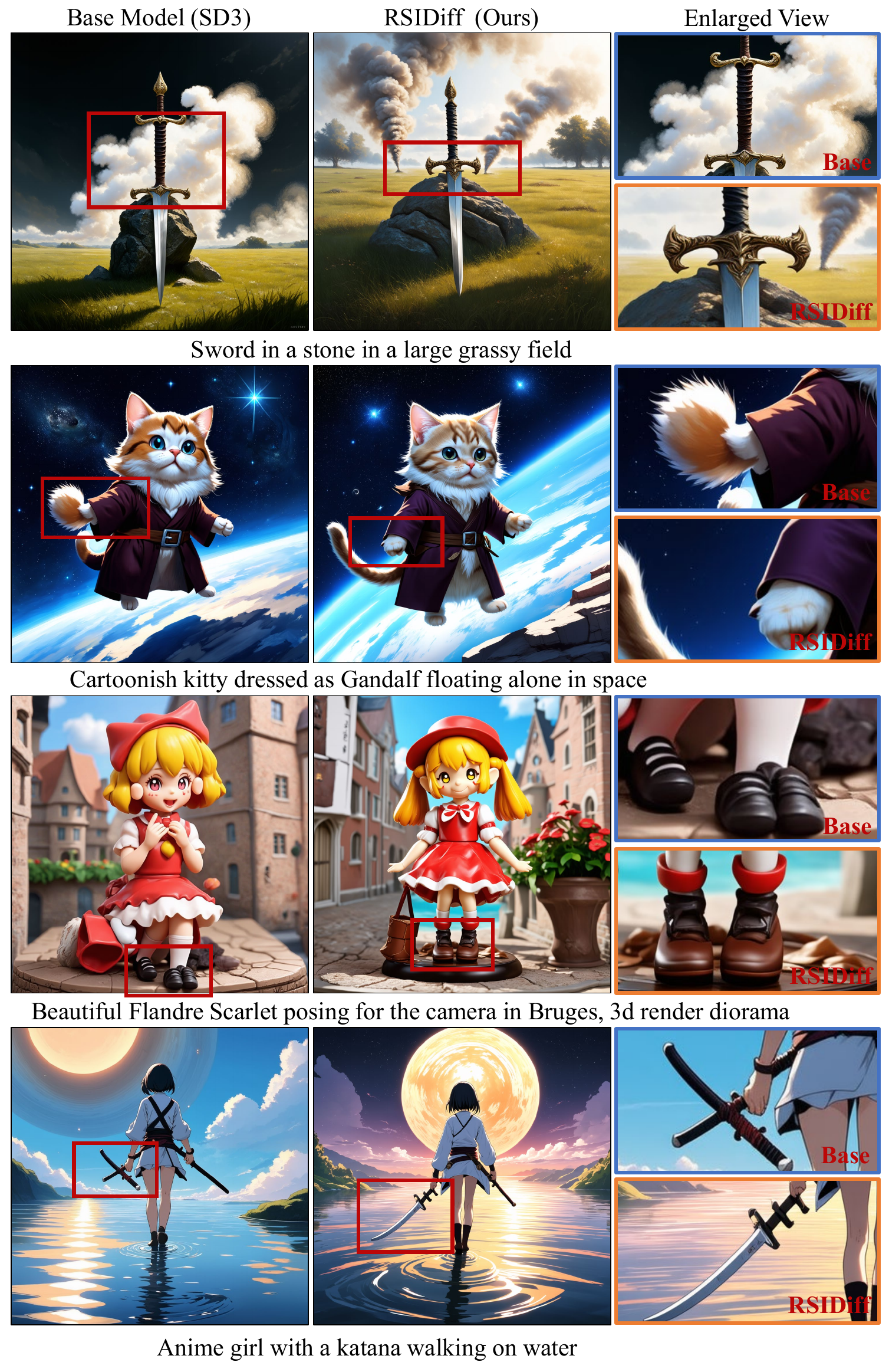}}
\caption{\textbf{Examples Generated by SD3 and RSIDiff.} This comparison highlights that our method RSIDiff enhances the base model by generating physically plausible subjects.}
\label{fg:supp_sd3_3}
\end{center}
\end{figure*}

We provide more qualitative results to show the effectiveness of RSIDiff.
\cref{fg:supp_iter} illustrates how RSIDiff improves human preference as training progresses.
\cref{fg:supp_sd_samples} shows the superior performance of RSIDiff through the comparison with base model and SFT method.
\cref{fg:supp_sd3_1}, \cref{fg:supp_sd3_2}, and \cref{fg:supp_sd3_3} show that RSIDiff is able to boost the SD3 model across multiple aspects.

\begin{table*}[t]
    \centering
    \small
    \begin{tabular}{|c|p{15cm}|}
        \hline
        No. & Prompt \\ \hline
        1&Red panda in a spacesuit in space having an epiphany nebula in the background, trending on artstation, highly detailed \\\hline
        2 &  A samoyed dog seated on a rock in a jungle, mist, tropical trees, vines, birds, sunset, fluffy clouds, warm colors, beautiful lighting, digital art, intricate details, trending on artstation\\\hline
        3&Digital art, fantasy portrait of a cat in a lounge chair wearing sunglasses, by james jean, by ross tran, ultra detailed, character design, concept art, trending on artstation\\\hline
         4& Illustration of a short curly orange hair man as a portrait, smooth, reflects, masterpiece artwork, ultra detailed, artgerm, style by karl marx, digital art, trending on artstation, behance, deviantart \\\hline
         5& Fennec fox, pink, palm trees, furry, cute, disney zootopia, concept art, aviator sunglasses, synthwave style, artstation, detailed, award winning, dramatic lighting, miami vice, oil on canvas\\\hline
         6&Digital art of a cute penguin sitting on a chair wearing sunglasses at night, detailed, trending on artstation, digital art, award winning art, detailed digital art, painting\\\hline
         7&A dog in an astronaut suit, 3d, sci-fi fantasy, intricate, elegant, highly detailed, lifelike, photorealistic, digital painting, artstation, illustration, concept art, sharp focus, art in the style of Shigenori Soejima \\\hline
         8&Portrait of a vampire woman staring into a mirror, realistic, 8 k, extremely detailed, cgi, trending on artstation, hyper - realistic render, 4 k hd wallpaper, premium prints available, by greg rutkowski\\\hline
        9&A medieval brick castle surrounded by dozens of planets in a green field at noon, matte oil painting, cumulus clouds, impact craters, fantasy, concept art, clear, crisp, sharp, extremely detailed, wallpaper\\\hline
        %
        %
        %
        %
    \end{tabular}
    \caption{\textbf{Sample Prompts from the Prompt Set.} This table displays several prompts included in our prompt set, which are selected based on three key aspects: clarity, specificity, and diversity.}
    \label{tb:prompt}
\end{table*}

\end{document}